\documentclass[journal]{IEEEtran}
\usepackage{tikz,xcolor}
\usepackage{colortbl}

\usepackage{amsmath,amsfonts}
\usepackage{algorithmic}
\usepackage[ruled,vlined]{algorithm2e} 
\usepackage{array}
\usepackage[caption=false,font=normalsize,labelfont=sf,textfont=sf]{subfig}
\usepackage{textcomp}
\usepackage{stfloats}
\usepackage{url}
\usepackage{verbatim}
\usepackage{graphicx}
\usepackage{cite}
\usepackage{multirow}
\usepackage{bm}
\usepackage{balance}
\usepackage{booktabs} 
\usepackage{xcolor}         
\usepackage{hyperref}
\hypersetup{
    colorlinks=true,     
    linkcolor=blue,      
    citecolor=blue,      
}

\definecolor{lime}{HTML}{A6CE39}
\DeclareRobustCommand{\orcidicon}{
\begin{tikzpicture}
\draw[lime, fill=lime] (0,0)
circle[radius=0.16]
node[white]{{\fontfamily{qag}\selectfont \tiny \.{I}D}};
\end{tikzpicture}
\hspace{-2mm}
}
\foreach \x in {A, ..., Z}{%
\expandafter\xdef\csname orcid\x\endcsname{\noexpand\href{https://orcid.org/\csname orcidauthor\x\endcsname}{\noexpand\orcidicon}}
}

\begin{document}

\title{PriorCLIP: Visual Prior Guided Vision-Language Model for Remote Sensing Image-Text Retrieval}

\author{
Jiancheng Pan\hspace{-1.5mm}\orcidA{},~\IEEEmembership{Student Member,~IEEE,}
Muyuan Ma, 
Qing Ma\hspace{-1.5mm}\orcidD{}, ~\IEEEmembership{Member,~IEEE,}\\
Cong Bai\hspace{-1.5mm}\orcidB{},~\IEEEmembership{Member,~IEEE,}
and Shengyong Chen\hspace{-1.5mm}\orcidC{},~\IEEEmembership{Senior Member,~IEEE,}

\thanks{Manuscript created October, 2020; This work was developed by the IEEE Publication Technology Department. This work is distributed under the \LaTeX \ Project Public License (LPPL) ( http://www.latex-project.org/ ) version 1.3. A copy of the LPPL, version 1.3, is included in the base \LaTeX \ documentation of all distributions of \LaTeX \ released 2003/12/01 or later. The opinions expressed here are entirely that of the author. No warranty is expressed or implied. User assumes all risk.}}

\markboth{Journal of \LaTeX\ Class Files,~Vol.~18, No.~9, September~2020}%
{How to Use the IEEEtran \LaTeX \ Templates}


\maketitle

\begin{abstract}
Remote sensing image-text retrieval plays a crucial role in remote sensing interpretation, yet remains challenging under both \emph{closed-domain} and \emph{open-domain} scenarios due to semantic noise and domain shifts. To address these issues, we propose a visual prior-guided vision-language model, \textbf{PriorCLIP}, which leverages visual priors for unbiased representation learning and adaptive vision-language alignment. In the closed-domain setting, PriorCLIP introduces two Progressive Attention Encoder (PAE) structures: Spatial-PAE constructs a belief matrix with instruction embeddings to filter key features and mitigate semantic bias. At the same time, Temporal-PAE exploits cyclic activation across time steps to enhance text representation. For the open-domain setting, we design a two-stage prior representation learning strategy, consisting of large-scale pre-training on coarse-grained image-text pairs, followed by fine-tuning on fine-grained pairs using vision-instruction, which enables robust retrieval across long-tail concepts and vocabulary shifts. Furthermore, a cluster-based symmetric contrastive Attribution Loss is proposed to constrain inter-class relations and alleviate semantic confusion in the shared embedding space. Extensive experiments on RSICD and RSITMD benchmarks demonstrate that PriorCLIP achieves substantial improvements, outperforming existing methods by $4.9\%$ and $4.0\%$ in closed-domain retrieval, and by $7.3\%$ and $9.4\%$ in open-domain retrieval, respectively.
\end{abstract}


\section{Introduction}
\IEEEPARstart{R}{emote} Sensing Image-Text Retrieval (RSITR) is a critical research task that utilizes image patches (or text queries) to retrieve their corresponding textual (or visual) counterparts from large-scale remote sensing databases acquired by satellites or aerial platforms, thereby facilitating the extraction of valuable semantic information \cite{8613838,9829420}. Owing to its significant role in practical applications such as natural resource exploration, disaster prevention, and environmental monitoring \cite{9296550,9474911,9696168,chi2016big}, RSITR has attracted increasing attention from both academia and industry. In recent years, the explosive growth of remote sensing data \cite{lu2017exploring} and the rapid development of cross-modal retrieval methodologies \cite{9429177,10138903} have jointly provided unprecedented opportunities, enabling RSITR to achieve enhanced efficiency, scalability, and accuracy.  

\begin{figure}[t]
    \centering
    \includegraphics[width=0.9\linewidth]{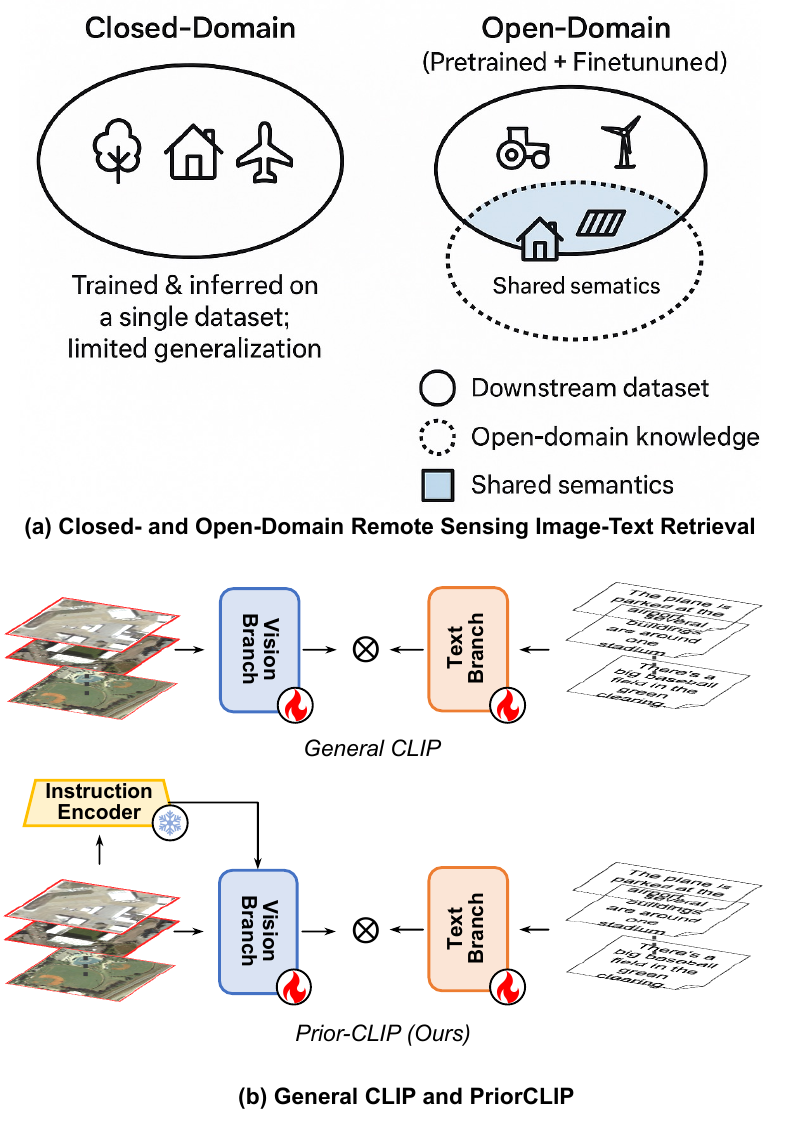}
    \caption{(a): Closed- and open-domain remote sensing image-text retrieval. (b): Unlike general CLIP structures, PriorCLIP leverages visual priors to guide the image-text alignment process.}
    \label{fig:fig1}
\end{figure}

\begin{figure*}[t]
    \centering
    \includegraphics[width=\linewidth]{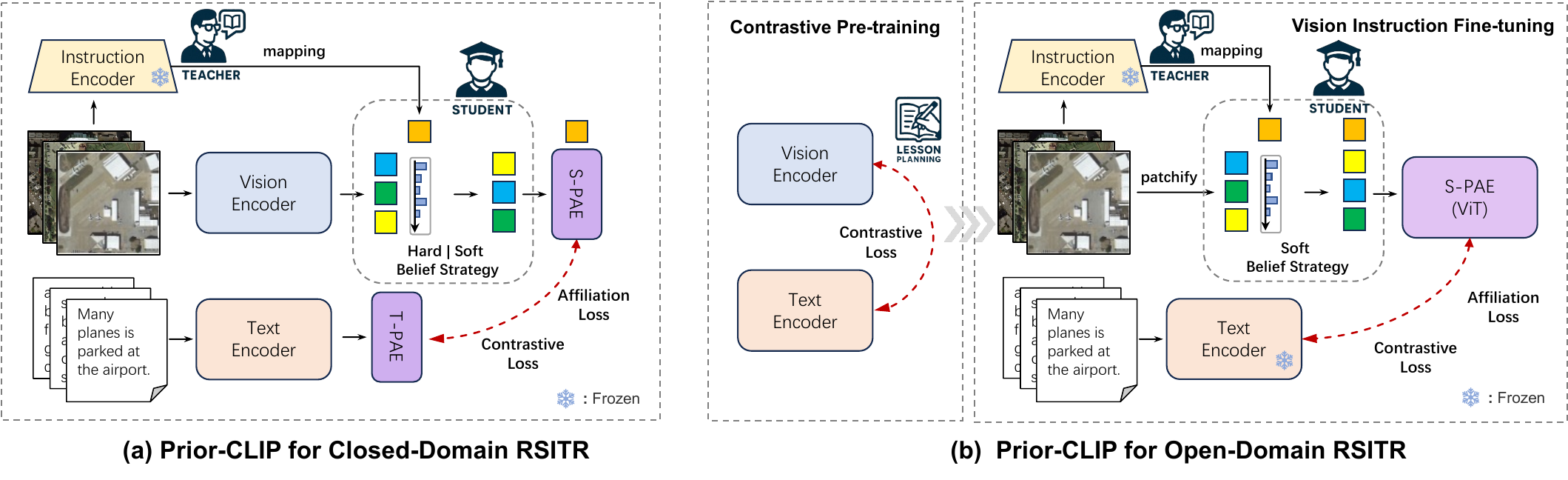}
    \caption{The PriorCLIP's structure. (a) PriorCLIP for closed-domain RSITR, which uses Spatial-PAE (S-PAE) and Temporal-PAE (T-PAE) for unbiased representation in the common subspace (\textcolor{blue}{Teacher instructs Student}); (b) PriorCLIP for open-domain RSITR, which uses a two-stage strategy to pre-train vision and text encoders on large-scale remote sensing image-text pairs, then perform vision instruction fine-tuning for open-domain retrieval (\textcolor{blue}{Teacher instructs Student through Lesson Planning}).}
    \label{fig:fig2}
\end{figure*}

\IEEEpubidadjcol
According to different training paradigms, RSITR can be divided into closed-domain and open-domain remote sensing image-text retrieval, as shown in Fig.~\ref{fig:fig1}(a). Closed-Domain RSITR (CD-RSITR) is a supervised training and retrieval method that operates on a single dataset; however, it is less capable of retrieving more complex ground elements. In closed-domain retrieval, most existing methods utilize CNN-based \cite{lecun1998gradient} vision representations and RNN-based \cite{hochreiter1997long} text representations, employing pair-wise triplet loss \cite{karpathy2015deep} for optimization. Abdullah et al. \cite{abdullah2020textrs} first applied modality fusion to solve the remote sensing image-text matching problem. Lv et al. \cite{lv2021fusion} utilized knowledge distillation to transform the fusion information of modalities into representations for vision and text learning. Mi et al. \cite{mi2022knowledge} utilized a knowledge graph to enrich text semantics and to reduce the information gap between vision and language. Yuan et al. \cite{yuan2022remote} developed an image-text retrieval model to understand multi-level visual semantics, leveraging both global and local information. Pan et al. \cite{pan2023reducing} proposed a scene-aware aggregation network to improve the fine-grained perception of the scene. These approaches benefit from a clear vision and text representation, but perform poorly for long-range dependency modeling \cite{dieng2016topicrnn, cao2019gcnet}. Open-Domain RSITR (OD-RSITR) is fine-tuned on a small dataset after pre-training on an additional large-scale dataset, resulting in improved ground element recognition. CLIP \cite{radford2021learning} inspired efforts with powerful zero-shot learning on visual interpretation tasks in open-domain retrieval. Subsequently, RemoteCLIP \cite{remoteclip}, GeoRSCLIP \cite{10679571}, and SkyCLIP \cite{Wang_Prabha_Huang_Wu_Rajagopal_2024} models were introduced to establish a foundation vision-language model for remote sensing (RSVLM) and improve large-scale open-domain image-text retrieval performance. Their objective is to develop a foundation of RSVLM, rather than exploring methods to enhance the visual representation of remote sensing.

Despite these advances, RSITR faces challenges that are distinct from those in natural image-text retrieval. Remote sensing images often contain small-scale objects that are more susceptible to semantic noise \cite{yuan2022remote}, such as cluttered backgrounds or irrelevant objects. Excessive sensitivity to such noise leads to biased representations and semantic confusion zones \cite{pan2023reducing}, i.e., overlapping regions in the shared embedding space that degrade retrieval accuracy. Efforts to address this problem include redundant feature filtering \cite{yuan2022exploring} and scene-aware aggregation \cite{pan2023reducing}. However, most approaches still rely on CNN- and RNN-based encoders, which are ineffective for capturing long-range dependencies and scaling with growing data volumes \cite{liang2021swinir}. With the emergence of Transformer-based models in computer vision \cite{dosovitskiy2020image} and natural language processing \cite{devlin2018bert}, Transformer encoders for vision and text have shown promise in RSITR. For example, Zhang et al. \cite{zhang2022transformer} designed a Transformer-based decoupling module to obtain invariant features across modalities. However, these methods often neglect the unique properties of remote sensing imagery \cite{yuan2022remote, pan2023reducing}, resulting in limited performance gains. Therefore, a fundamental question arises: \textit{how can we design an RSVLM tailored to the specific challenges of RSITR?}

In this paper, we propose a visual prior-guided vision-language model, \textbf{PriorCLIP}, which leverages the visual prior of remote sensing scene recognition \cite{zou2015deep} to guide the image and text alignment as shown in Fig. \ref{fig:fig1}(b). For closed-domain image-text retrieval, two Progressive Attention Encoder (PAE) structures, Spatial-PAE and Temporal-PAE, are proposed to perform long-range dependency modelling to enhance key feature representation as shown in Fig. \ref{fig:fig2}(a).
Spatial-PAE leverages instruction embeddings to construct a \emph{belief matrix} that softly filters visual tokens, yielding an unbiased visual representation by reweighting reliable regions and suppressing spurious cues; Temporal-PAE propagates information along the textual sequence by using the previous time step to activate the current step, thereby improving the quality and temporal consistency of text representations. For the open-domain setting, we propose a two-stage prior representation learning pipeline that targets semantic noise in remote sensing vision--language alignment: (i) large-scale pre-training of the image and text encoders on \(\sim\!5\) million \emph{coarse-grained} image--text pairs to acquire broad coverage of the remote sensing vocabulary and scene compositions; followed by (ii) \emph{vision instruction} fine-tuning on \emph{fine-grained} pairs to sharpen concept boundaries and enhance compositional generalization. To further regularize the shared embedding space, we devise a \emph{cluster-based symmetric contrastive Attribution Loss} that jointly pulls attribution-consistent positives while enlarging inter-class margins from both the image and text sides, effectively shrinking semantic confusion zones in the common subspace. Extensive experiments demonstrate consistent gains on mAP for both closed- and open-domain retrieval, with ablations validating the contributions of Spatial-/Temporal-PAE, belief-based filtering, the two-stage prior learning strategy, and the proposed attribution loss.
Our PriorCLIP achieves improvements in the closed-domain RSITR of 4.9\% and 4.0\% on RSICD~\cite{lu2017exploring} and RSITMD~\cite{yuan2022exploring}, respectively, and outperforms open-domain methods by 7.3\% and 9.4\%. Our key contributions are as follows:
\begin{itemize}
\item We introduce the visual prior-guided vision-language model, \textbf{PriorCLIP}, to enhance remote sensing scene understanding by leveraging visual priors for unbiased representation and enabling both closed-domain and open-domain retrieval.
\item For closed-domain retrieval, two PAE structures, Spatial-PAE and Temporal-PAE, are proposed to perform long-range dependency modelling to enhance key feature representation. We use instruction embedding to filter features by building a belief matrix, aiming for an unbiased vision representation. Temporal-PAE improves text representation by using the previous time step to activate the current time step.
\item For open-domain retrieval, a two-stage method with prior representation learning is proposed to address semantic noise in remote sensing vision-language representations and further enhance open-domain retrieval performance. First, the image and text encoders are pre-trained on five million coarse-grained remote sensing image-text pairs, and then fine-tuning is performed on vision instructions using fine-grained image-text pairs.
\item We propose a cluster-based symmetric contrastive Attribution Loss to constrain inter-class distances and reduce semantic confusion zones in the common subspace.
\end{itemize}

\noindent Compared with our previous work~\cite{pan2023prior} (ACMMM~2023), this paper advances RSITR with three points. We first provide a principled synthesis of \emph{closed-domain} vs. \emph{open-domain} RSITR, formalizing problem settings and assumptions on label space and domain shift, unifying representative methods, and clarifying evaluation criteria (both closed-domain and open-domain). Building on this, we extend the visual prior-guided vision-language model framework to genuinely open-domain retrieval via lightweight, prior-guided adapters that interface with large vision-language backbones, thereby supporting compositional queries and long-tail concepts with minimal task-specific tuning, improved robustness to vocabulary shifts, and efficient inference. Finally, we enhance PriorCLIP by introducing a soft-belief refinement mechanism that assigns continuous belief weights to visual tokens, emphasizing reliable regions while suppressing noisy cues. This effectively mitigates semantic noise and reduces spurious correlations, with benefits particularly pronounced in open-domain scenarios. Extensive experiments across standard RSITR benchmarks confirm consistent gains, with comprehensive ablations and closed-domain and open-domain retrieval studies validating the effectiveness of the proposed PriorCLIP.

\section{RELATED WORK}
\subsection{Remote Sensing Image-text Retrieval}
Although traditional image-text retrieval methods based on natural images are increasingly mature, they cannot be directly applied to multiscale and redundant remote sensing images. There is still a long way to go in Remote Sensing Image-Text Retrieval (RSITR). According to different training paradigms, RSITR methods can be divided into closed-domain RSITR (CD-RSITR) and open-domain RSITR (OD-RSITR) methods.

\textbf{Closed-Domain method} is a supervised method for training and retrieval under a single dataset,  but it is less capable of retrieving more complex ground elements. Some works \cite{mao2018deep,abdullah2020textrs,mi2022knowledge,yuan2022remote,zhang2022transformer} only perform self-interaction on the same modality to obtain an enhanced modal semantic representation. \cite{mao2018deep,abdullah2020textrs} used the information of different modalities to obtain a joint semantic representation. Might et al. \cite{mi2022knowledge} proposed a knowledge-based method to solve the problem of coarse-grained textual descriptions. Yuan et al. \cite{yuan2022remote} proposed a network structure that combines global and local information, utilizing attention-based modules to fuse multi-level information dynamically. Zhang et al. \cite{zhang2022transformer} designed a reconstruction module to reconstruct the decoupled features, ensuring the maximum retention of information in these features. Some others \cite{lv2021fusion,yuan2022mcrn,cheng2021deep,yuan2022exploring} use information between different attention mechanisms or shared parameter networks for interactive learning. Lv et al. \cite{lv2021fusion} propose a fusion-based association learning model to fuse image and text information across modalities to improve the semantic relevance of different modalities. Specifically, Yuan et al. \cite{yuan2022mcrn} employed a shared pattern transfer module to facilitate interaction between modalities, thereby addressing the semantic heterogeneity issue between different modal data. Cheng et al. \cite{cheng2021deep} propose a semantic alignment module that uses an attention mechanism to enhance the correspondence between image and text. Yuan et al. \cite{yuan2022exploring} suggest an asymmetric multimodal feature matching network to adapt multiscale feature inputs while using visual features to guide text representation. These methods are based on partial simulations of the real, complex remote sensing world and fail to learn fine-grained perceptions of the environment.

\textbf{Open-Domain method} is fine-tuned on a small dataset after pre-training on an additional large-scale dataset, with better ground element recognition. Since the advent of CLIP \cite{radford2021learning}, CLIPs for tasks in remote sensing \cite{remoteclip, 10679571, Wang_Prabha_Huang_Wu_Rajagopal_2024, djoufack2022clip} have emerged. Some work \cite{djoufack2022clip} has begun early exploration of fine-tuning CLIP to enhance remote sensing image-text retrieval, but it has only achieved poor performance. Further, Chen et al. \cite{remoteclip} proposed RemoteCLIP, the first vision-language model for remote sensing, to implement language-guided retrieval and zero-shot tasks. Zhang et al. \cite{10679571} published RS5M, a large-scale remote sensing image-text dataset with more than 5 million images, and utilized geographic information to train the vision-language model GeoRSCLIP. Coincidentally, a new remote sensing benchmark dataset, SkyScript, and a benchmark model, SkyCLIP, were proposed by Wang \cite{Wang_Prabha_Huang_Wu_Rajagopal_2024} et al. However, their goal is mainly to build a foundation for a vision-language model for remote sensing (RSVLM), without further exploring how to enhance the visual representation of remote sensing. We aim to propose a CLIP-based model that enhances the performance of large-scale, open-domain remote sensing retrieval, building upon the foundation of RSVLM.

\subsection{Transformer-based Cross-Attention Mechanism}
Different from self-attention~\cite{vaswani2017attention}, which models dependencies within a single sequence, cross-attention is a mechanism in the Transformer architecture that integrates information across two distinct embedding sequences, often originating from different modalities or outputs~\cite{gheini2021strengths}. Due to its ability to capture cross-modal interactions, cross-attention has been widely adopted in various tasks, such as image-text retrieval~\cite{lee2018stacked,wei2020multi,xu2020cross,10105185,10138903} and recommendation~\cite{9837877,9881215,10143372}.  

In image-text retrieval, Lee et al.~\cite{lee2018stacked} proposed the \textit{Stacked Cross Attention} model, which enables contextual alignment between images and sentences. Wei et al.~\cite{wei2020multi} introduced the \textit{MultiModality Cross Attention} Network to jointly capture intra- and inter-modal relationships in a unified framework. Xu et al.~\cite{xu2020cross} developed the CASC framework, which combines cross-modal attention for local alignment with multilabel prediction to enforce global semantic consistency. However, these approaches are often tailored to specific tasks and lack general applicability. To address this, Diao et al.~\cite{10105185} designed two plug-and-play regulators for adaptive contextualization and aggregation of cross-modal features. Liu et al.~\cite{10138903} further proposed a bidirectional correct attention network that adjusts attention weights by modeling the relevance between sub-fragments and global semantics. Despite their effectiveness, most existing methods struggle to scale with large temporal- and spatial-scale data, highlighting the need for a universal and easily adaptable cross-attention structure.

\section{Remote sensing image-text retrieval}
\subsection{Generic Image and Text Encoding}

A common approach to interpreting remote sensing images~\cite{9167483} is to align visual and textual information at the semantic level. Remote Sensing Image-Text Retrieval (RSITR) serves as a fundamental task for this alignment by encoding image and text features into a shared semantic space.  

For image representation, existing RSITR methods employ either CNN-based encoders~\cite{abdullah2020textrs,cheng2021deep,lv2021fusion,mi2022knowledge,yuan2022mcrn,yuan2021lightweight,yuan2022exploring,yuan2022remote} or Transformer-based encoders~\cite{bai2018optimization,zhang2023omcbir}.  
Formally, given an RGB image 
$\mathbf{I}_{\mathrm{img}} \in \mathbb{R}^{3 \times H \times W}$,  
an image encoder $\mathbf{E}_{\mathrm{img}}$ maps it into a global feature $\mathbf{f}_{\mathrm{cls}} \in \mathbb{R}^{d}$ (corresponding to \texttt{[CLS]}) and a set of local features $\mathbf{F}_v \in \mathbb{R}^{d \times m}$:  
\begin{equation}
\bigl[\mathbf{f}_{\mathrm{cls}},\, \mathbf{F}_v \bigr] 
= \mathbf{E}_{\mathrm{img}}\!\left(\mathbf{I}_{\mathrm{img}};\, \Theta_{\mathrm{img}}\right),
\label{eq:image_encoding}
\end{equation}
where $\Theta_{\mathrm{img}}$ denotes the trainable parameters of the encoder, and $[\cdot,\cdot]$ indicates concatenation along the sequence dimension. For CNN-based encoders, $\mathbf{f}_{\mathrm{cls}}$ captures high-level semantics, while $\mathbf{F}_v$ corresponds to spatially aggregated feature maps.  

For text representation, RNN-based encoders~\cite{abdullah2020textrs,yuan2022exploring,yuan2022remote} and Transformer-based encoders~\cite{mi2022knowledge,zhang2022transformer} are most commonly used. Compared with RNNs~\cite{chung2014empirical}, Transformers~\cite{vaswani2017attention} capture global semantic dependencies by attending to all positions in the sequence simultaneously.  
Given a text sequence 
$\mathbf{T}_{\mathrm{txt}} = \{w_1, w_2, \dots, w_n\}$,  
the text encoder $\mathbf{E}_{\mathrm{txt}}$ yields a global feature $\mathbf{t}_{\mathrm{cls}} \in \mathbb{R}^{d}$ and local token embeddings $\mathbf{F}_t \in \mathbb{R}^{d \times n}$:  
\begin{equation}
\bigl[\mathbf{t}_{\mathrm{cls}},\, \mathbf{F}_t \bigr] 
= \mathbf{E}_{\mathrm{txt}}\!\left(\mathbf{T}_{\mathrm{txt}};\, \Theta_{\mathrm{txt}}\right),
\label{eq:text_encoding}
\end{equation}
where $\Theta_{\mathrm{txt}}$ denotes the trainable parameters of the text encoder, and $n$ is the number of tokens in the input sequence. In RNN-based encoders, $\mathbf{t}_{\mathrm{cls}}$ is often obtained by applying a pooling operation over $\mathbf{F}_t$.  

In summary, both $\bigl[\mathbf{f}_{\mathrm{cls}}, \mathbf{F}_v \bigr]$ from Eq.~\eqref{eq:image_encoding} and $\bigl[\mathbf{t}_{\mathrm{cls}}, \mathbf{F}_t \bigr]$ from Eq.~\eqref{eq:text_encoding} provide the basis for subsequent cross-modal alignment, while the specific extraction strategies vary depending on encoder architectures.

\begin{figure}[t]
    \centering
    \includegraphics[width=0.85\linewidth]{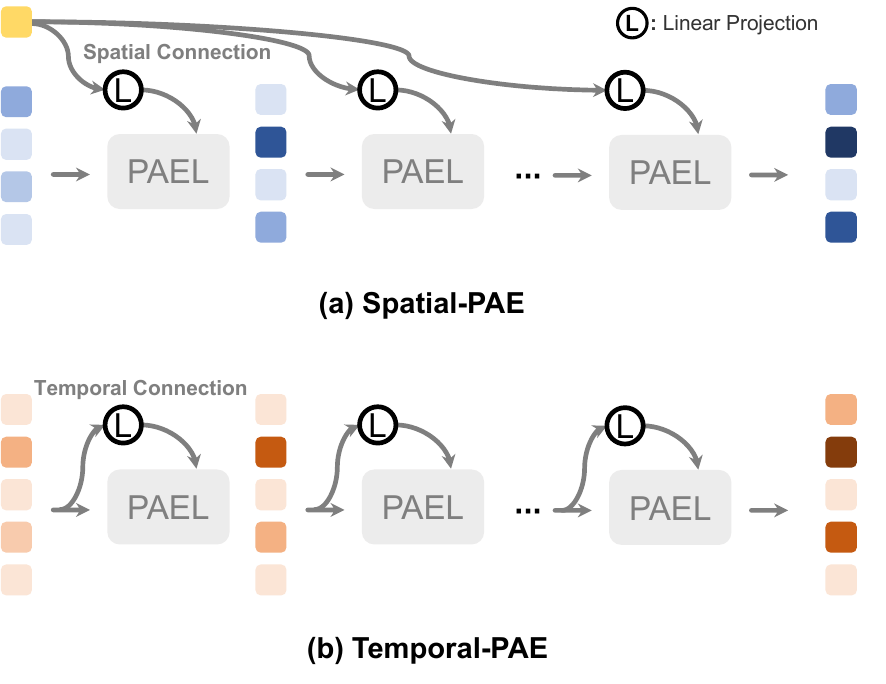}
    \caption{Two different progressive attention encoders (PAE) for messaging: (a) Spatial-PAE and (b) Temporal-PAE.}
    \label{fig:fig4}
\end{figure}

\subsection{Various Image-Text Alignment}

A straightforward approach to image-text alignment is to fine-tune the vision and text encoders with learnable weights $\Theta_{\mathrm{img}}$ and $\Theta_{\mathrm{txt}}$, aligning the global features $\mathbf{f}_{\mathrm{cls}}$ and $\mathbf{t}_{\mathrm{cls}}$ directly, as in VSE~\cite{faghri2017vse++}. This method performs well on datasets with abundant paired samples (e.g., COCO~\cite{lin2014microsoft}), but its improvement is limited in remote sensing scenarios.  

To capture richer semantics, several works such as SCAN~\cite{yuan2021lightweight}, CAMP~\cite{lee2018stacked}, and CAMERA~\cite{qu2020context} introduce feature-level interactions between image and text representations. This can be formalized as  
\begin{equation}
\mathbf{Z}_{\mathrm{img\!-\!txt}} = \mathbf{E}_{\mathrm{img\!-\!txt}}\!\left(\mathbf{F}_v,\, \mathbf{F}_t;\, \Theta_{\mathrm{img\!-\!txt}}\right),
\label{eq:img_txt_interaction}
\end{equation}
where $\mathbf{E}_{\mathrm{img\!-\!txt}}$ denotes a cross-modal interaction module parameterized by $\Theta_{\mathrm{img\!-\!txt}}$. While effective in natural image retrieval, these methods face challenges in remote sensing due to variable imaging scales and the difficulty of fine-grained alignment.  

In most RSITR approaches, visual information is further refined since remote sensing images often contain substantial redundancy. For instance, AMFMN~\cite{yuan2022exploring} and SWAN~\cite{pan2023reducing} perform multi-scale fusion of $\mathbf{f}_{\mathrm{cls}}$ and $\mathbf{F}_v$, filtering redundant features to highlight salient visual cues. GaLR~

\section{Visual Prior Guided Vision-Language Model}
\subsection{Instruction Encoding}

Remote sensing images often suffer from semantic noise, where background clutter or irrelevant objects bias the visual representation. Recent studies~\cite{li2018guiding,dathathri2019plug,kirillov2023segment,yang2023track,wang2023caption,10158512} have shown that introducing prompts or instructions can effectively guide models toward more reliable representations. Inspired by this idea, we adopt a ResNet~\cite{he2016deep} pre-trained on the AID dataset~\cite{xia2017aid} as an instruction encoder
\begin{equation}
\mathbf{f}_{\mathrm{ins}} = \mathbf{E}_{\mathrm{ins}}\!\left(\mathbf{I}_{\mathrm{img}};\, \Theta_{\mathrm{ins}}\right),
\label{eq:instruction_encoding}
\end{equation}
where $\Theta_{\mathrm{ins}}$ denotes the pre-trained weights. The encoder produces an instruction embedding $\mathbf{f}_{\mathrm{ins}} \in \mathbb{R}^{d}$, which provides prior knowledge for refining visual features.  

Unlike GaLR~\cite{yuan2022remote}, which incorporates $\mathbf{F}_v$ with external object-level knowledge via GCN and fuses them directly—potentially disturbing intrinsic image semantics—our approach treats instruction embeddings as \emph{guidance} rather than as additional features. Specifically, the instruction embedding $\mathbf{f}_{\mathrm{ins}}$ is used to filter and reweight the raw visual features $\bigl[\mathbf{f}_{\mathrm{cls}}, \mathbf{F}_v\bigr]$ through an image–instruction interaction module:  
\begin{equation}
\mathbf{Z}_{\mathrm{img\!-\!ins}} = \mathbf{E}_{\mathrm{img\!-\!img}}\!\left(\bigl[\mathbf{f}_{\mathrm{cls}}, \mathbf{F}_v\bigr];\, \mathbf{f}_{\mathrm{ins}},\, \Theta_{\mathrm{img\!-\!img}}\right).
\label{eq:img_instruction_interaction}
\end{equation}

In this way, the model selectively emphasizes unbiased and semantically reliable visual features, reducing noise and improving the quality of visual representations.

\subsection{Hard and Soft Belief Strategy}\label{Se3.3}

Remote sensing images often contain small-scale objects that are easily disturbed by background clutter or irrelevant regions, which amplifies semantic noise in visual representations. To mitigate this issue, we design a belief strategy that leverages instruction embeddings to construct a belief matrix, guiding the filtering of visual features toward an unbiased representation, as illustrated in Fig.~\ref{fig:fig2}.  

\textbf{Hard Belief Strategy.}  
Given prior knowledge from remote sensing scene recognition $\mathbf{f}_{\mathrm{ins}} \in \mathbb{R}^{d}$, we compute a belief matrix $\mathbf{M}^{\text{bel}} \in \mathbb{R}^{1 \times (m+1)}$ over the concatenated features $\bigl[\mathbf{f}_{\mathrm{cls}}, \mathbf{F}_v\bigr]$:  
\begin{equation}
\mathbf{M}^{\text{bel}}
= \operatorname{Softmax}\!\left(\mathbf{f}_{\mathrm{ins}}^{\mathrm{T}} \bigl[\mathbf{f}_{\mathrm{cls}},\, \mathbf{F}_v\bigr]\right).
\label{eq:belief_matrix}
\end{equation}
Features are then sorted and filtered according to $\mathbf{M}^{\text{bel}}$ to retain the top-$k$ visual tokens:  
\begin{equation}
\bigl[\mathbf{f}_{\mathrm{cls}}, \mathbf{F}_v\bigr] 
\underset{\text{Sort \& Filter}}{\stackrel{\mathbf{M}^{\text{bel}}}{\longrightarrow}} 
\mathbf{F}_v^{(k)}=\bigl[\mathbf{F}_{:,r_1}, \ldots, \mathbf{F}_{:,r_k}\bigr], 
\quad (r_k < m+1).
\label{eq:hard_filter}
\end{equation}
This hard filtering enforces discrete feature selection but requires a manually set $k$, making it sensitive to empirical choices.  

\textbf{Soft Belief Strategy.}  
To avoid fixed feature selection, we propose a soft belief strategy that dynamically reweights features instead of discarding them. For each element in $\mathbf{M}^{\text{bel}}$, we compute its batch-wise rank $R_j$:  
\begin{equation}
R_j = 1 + \sum_{k=1}^{m+1} \mathbb{I}\!\left(\mathbf{M}^{\text{bel}}_{k}<\mathbf{M}^{\text{bel}}_{j}\right),
\label{eq:rank}
\end{equation}
where $\mathbb{I}(\cdot)$ is the indicator function. Each visual token $\mathbf{F}_{:,l} \in \mathbb{R}^{d}$ is then reweighted based on both its belief score and relative rank:  
\begin{equation}
\mathbf{F}_v^{(\text{soft})}= \sum_{l=1}^{m+1} \mathbf{F}_{:,l} 
\left(\mathbf{M}^{\text{bel}}_{l} + \frac{1}{\sqrt{R_l}}\right).
\label{eq:soft_belief}
\end{equation}
This produces rank-weighted features that preserve all tokens while emphasizing reliable ones. Unlike Eq.~\eqref{eq:hard_filter}, this strategy requires no predefined $k$, offering better adaptability.  

\textbf{Discussion.}  
The soft belief strategy improves both retrieval accuracy and efficiency, as validated in Tab.~\ref{tab:table1}. Moreover, its dynamic filtering capability makes it especially suitable for open-domain retrieval, where large-scale training data and diverse concepts demand unbiased and flexible visual representations.

\subsection{Visual Prior Guided Image-Text Alignment}
\subsubsection{\textbf{Progressive Attention Encoder Layer}}
The Transformer Encoder Layer (TEL) is a fundamental component of the Transformer architecture and has been widely applied in both natural language processing and computer vision tasks~\cite{10479175,10381779,9906456}. Building upon TEL, we design a Progressive Attention Encoder Layer (PAEL) that integrates both self-attention~\cite{vaswani2017attention} and cross-attention~\cite{jaegle2021perceiver} to progressively enhance feature representations.  

Unlike self-attention, which models dependencies within a single sequence, cross-attention enables interaction between two distinct sequences, regardless of modality. Given two inputs $\mathbf{H}_{l-1}^S \in \mathbb{R}^{d \times N}$ and $\mathbf{H}_{l-1}^C \in \mathbb{R}^{d \times N}$, PAEL is computed as:  
\begin{align}
\mathbf{H}_{l+1}^S &= \mathrm{FFN}\bigl(\mathrm{Attn}(\mathbf{H}_{l-1}^S)\bigr), \label{eq:s-branch} \\
\mathbf{H}_{l+1}^C &= \mathrm{FFN}\bigl(\mathrm{Attn}(\mathbf{H}_{l-1}^C, \mathbf{H}_{l+1}^S)\bigr), \label{eq:c-branch}
\end{align}
where $\mathrm{Attn}(\cdot)$ denotes multi-head attention and $\mathrm{FFN}(\cdot)$ represents a feedforward multilayer perceptron.  

In essence, PAEL first strengthens the self-representation of sequence $\mathbf{H}_{l-1}^S$ through self-attention (Eq.~\ref{eq:s-branch}), and then uses the refined $\mathbf{H}_{l+1}^S$ as contextual guidance for updating $\mathbf{H}_{l-1}^C$ via cross-attention (Eq.~\ref{eq:c-branch}). This progressive mechanism allows information to be propagated step by step, enhancing both intra-sequence dependencies and inter-sequence interactions. For convenience, the PAEL computation can be compactly expressed as:  
\begin{equation}
PAEL\left(\mathbf{H}_{l-1}^S, \mathbf{H}_{l-1}^C\right).
\end{equation}  

Compared with parallel attention designs, PAEL adopts a progressive update strategy to ensure that cross-attention operates on refined and semantically enriched features. This not only reduces the risk of propagating noisy information across modalities but also facilitates more stable and discriminative feature learning.

\begin{algorithm}[t]
\caption{Cluster-based Symmetric Contrastive Affiliation Loss in a Mini-batch}
\small
\label{alg:batch_cluster_loss}
\KwIn{
    Image embeddings $\mathbf{I}_e \in \mathbb{R}^{B \times d_e}$, \\
    Text embeddings $\mathbf{T}_e \in \mathbb{R}^{B \times d_e}$, \\
    Labels $\mathbf{c} = (c_1,\ldots,c_B)^\top \in \{1,\ldots,C\}^B$, \\
    Temperature $t \in \mathbb{R}$, \\
    Small constant $\varepsilon > 0$
}
\KwOut{Symmetric contrastive loss $\mathcal{L}$}
\BlankLine

\Begin{
    \tcp{Step 1: $\ell_2$ normalization}
    $\tilde{\mathbf{I}}_e \gets \operatorname{diag}\!\bigl(\|\mathbf{I}_{e,i}\|_2^{-1}\bigr)\mathbf{I}_e$ \;
    $\tilde{\mathbf{T}}_e \gets \operatorname{diag}\!\bigl(\|\mathbf{T}_{e,i}\|_2^{-1}\bigr)\mathbf{T}_e$ \;

    \tcp{Step 2: One-hot label matrix and class counts}
    Construct $\mathbf{Y} \in \{0,1\}^{B \times C}$ with $Y_{ik} = \mathbb{1}\{c_i = k\}$ \;
    $\mathbf{s} \gets \mathbf{Y}^\top \mathbf{1}_B + \varepsilon \mathbf{1}_C$ \;

    \tcp{Step 3: Class prototypes in the batch}
    $\mathbf{I}_{\mathrm{cls}} \gets \operatorname{diag}(\mathbf{s})^{-1}\mathbf{Y}^\top \tilde{\mathbf{I}}_e$ \;
    $\mathbf{T}_{\mathrm{cls}} \gets \operatorname{diag}(\mathbf{s})^{-1}\mathbf{Y}^\top \tilde{\mathbf{T}}_e$ \;

    \tcp{Step 4: Sample-aligned centers}
    $\mathbf{I}_c \gets \mathbf{Y}\mathbf{I}_{\mathrm{cls}}$ \;
    $\mathbf{T}_c \gets \mathbf{Y}\mathbf{T}_{\mathrm{cls}}$ \;

    \tcp{Step 5: Scaled pairwise cosine similarity}
    $\mathbf{Z}_{I2T} \gets e^t \tilde{\mathbf{I}}_e \mathbf{T}_c^\top$ \;
    $\mathbf{Z}_{T2I} \gets e^t \tilde{\mathbf{T}}_e \mathbf{I}_c^\top$ \;

    \tcp{Step 6: Symmetric loss}
    Let $\mathbf{y} = (0,1,\ldots,B-1)^\top$ \;
    $\mathcal{L}_{I2T} \gets \mathrm{CE}(\mathbf{Z}_{I2T}, \mathbf{y})$ \;
    $\mathcal{L}_{T2I} \gets \mathrm{CE}(\mathbf{Z}_{T2I}, \mathbf{y})$ \;
    $\mathcal{L} \gets \tfrac{1}{2}\bigl(\mathcal{L}_{I2T} + \mathcal{L}_{T2I}\bigr)$ \;

    \Return $\mathcal{L}$ \;
}
\end{algorithm}

\subsubsection{\textbf{Progressive Attention Encoder}}

To effectively capture key feature representations, we design two complementary message transfer mechanisms based on PAEL: Spatial-PAE (S-PAE) and Temporal-PAE (T-PAE), as illustrated in Fig.~\ref{fig:fig4}.  

\textbf{Spatial-PAE.}  
Spatial-PAE (Fig.~\ref{fig:fig4}(a)) establishes spatial connections between the input sequence and an external source through linear projection, where the interaction is realized by cross-attention rather than direct feature fusion~\cite{pan2023reducing,ma2023direction,yuan2022remote}. In conventional PAELs, self-attention alone struggles to capture global information, since long-range dependencies require extended sequence distances to be connected. To overcome this limitation, Spatial-PAE leverages external guidance to model long-range dependencies on filtered features activated by prior information:  
\begin{align}
\mathbf{F}_v^{i} 
&= \operatorname{PAEL}\!\left( \mathbf{F}_v^{\,i-1},\, \mathbf{W}_s^{\,i}\,\mathbf{F}_{\mathrm{ins}} \right), \notag\\
&\quad i = 1,2,\dots,n_v, \qquad \mathbf{F}_v^{0}=\mathbf{F}_v,
\label{eq:spa_pae}
\end{align}
\noindent where $\mathbf{F}_{\mathrm{ins}}=\bigl[\mathbf{f}_{\mathrm{ins}}, \mathbf{f}_{\mathrm{ins}}, \ldots, \mathbf{f}_{\mathrm{ins}}\bigr] \in \mathbb{R}^{d \times r_k}$ denotes replicated instruction embeddings, $\mathbf{W}_s^{\,i} \in \mathbb{R}^{d \times d}$ is the $i$-th projection weight, and we write $\mathbf{F}_v^{i}=\bigl[\mathbf{f}_{r_1}^{\,i}, \ldots, \mathbf{f}_{r_k}^{\,i}\bigr]$ for the token-wise output at layer $i$. From the last PAEL, we obtain an unbiased local-correlated embedding as  
\begin{equation}
\mathbf{f}_{\mathrm{loc}} 
= \operatorname{Head}\!\left(\mathbf{f}_{r_1}^{\,n_v}\right),
\label{eq:local_embedding}
\end{equation}
where $\operatorname{Head}(\cdot)$ maps the head token to an unbiased representation. Finally, to preserve global semantic correlations, the overall visual embedding is derived by combining global and local information:  
\begin{equation}
\mathbf{v}_{\mathrm{emb}} = \mathbf{f}_{\mathrm{cls}} + \mathbf{f}_{\mathrm{loc}}.
\label{eq:visual_embedding}
\end{equation}  

Spatial-PAE enables global dependency modeling under the guidance of instruction embeddings, producing unbiased and semantically enriched visual representations that complement the standard self-attention pathway.

\begin{figure}[t]
    \centering
    \includegraphics[width=0.9\linewidth]{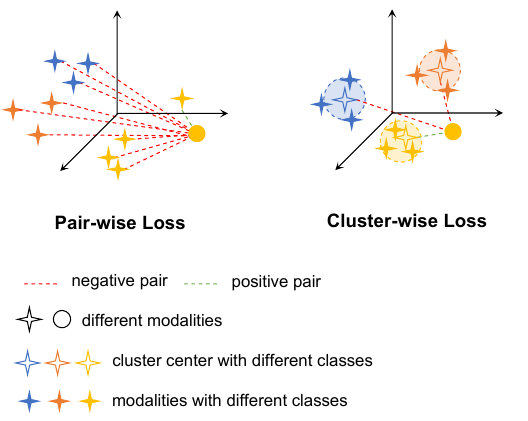}
    \caption{Pair-wise loss and cluster-wise loss.}
    \label{fig:fig5}
\end{figure}

\textbf{Temporal-PAE.}  
Temporal-PAE (Fig.~\ref{fig:fig4}(b)) establishes temporal connections across successive time steps through linear projection. While positional encoding is commonly used to capture sequential order, it may introduce ambiguity or inaccuracies when applied to deeper self-attention layers. To address this limitation, Temporal-PAE computes attention maps jointly from the outputs of the previous and current time steps, thereby providing more accurate positional cues and activating informative textual features.  

Formally, the recursive update of Temporal-PAE is defined as:  
\begin{equation}
\begin{split}
\mathbf{F}_t^{i} 
&= \operatorname{PAEL}\!\left(\mathbf{F}_t^{\,i-1},\, \mathbf{W}_t^{\,i}\mathbf{F}_t^{\,i-1}\right),\\
&\quad i = 1,2,\dots,n_t,\qquad \mathbf{F}_t^{0} = \bigl[\mathbf{t}_{\mathrm{cls}}, \mathbf{F}_t\bigr],
\end{split}
\label{eq:tmp_pae}
\end{equation}
where $\mathbf{W}_t^{\,i} \in \mathbb{R}^{d \times d}$ is the $i$-th projection weight, and $\mathbf{F}_t^{i}=\bigl[\mathbf{t}_{\mathrm{cls}}^{\,i}, \mathbf{e}_1^{\,i}, \mathbf{e}_2^{\,i}, \ldots, \mathbf{e}_n^{\,i}\bigr]$ denotes the token-wise output at layer $i$.  

From the last PAEL, an unbiased local-correlated embedding is obtained as  
\begin{equation}
\mathbf{t}_{\mathrm{loc}} 
= \operatorname{Head}\!\left(\mathbf{t}_{\mathrm{cls}}^{\,n_t}\right),
\label{eq:local_text_embedding}
\end{equation}
where $\operatorname{Head}(\cdot)$ maps the head token into a compact semantic representation. Finally, to preserve global semantic information, the overall text embedding is derived by combining global and local representations:  
\begin{equation}
\mathbf{t}_{\mathrm{emb}} = \mathbf{t}_{\mathrm{cls}} + \mathbf{t}_{\mathrm{loc}}.
\label{eq:text_embedding}
\end{equation}  

Temporal-PAE complements standard positional encoding by progressively propagating temporal dependencies across layers, enabling more reliable and unbiased text embeddings for retrieval tasks.

\subsubsection{\textbf{PriorCLIP for Closed-Domain and Open-Domain Retrieval}}
To address semantic noise and enhance cross-modal alignment, we extend PriorCLIP with a soft-belief strategy, improving upon the previous hard-belief mechanism~\cite{pan2023prior}. Specifically, we first collect positive remote sensing image–text pairs $\bigl\{(\mathbf{I}_{\mathrm{img}}^1, \mathbf{T}_{\mathrm{txt}}^1),\dots,(\mathbf{I}_{\mathrm{img}}^m, \mathbf{T}_{\mathrm{txt}}^m)\bigr\}$ from dataset $\mathcal{D}_{\mathrm{img\!-\!txt}}$. The image $\mathbf{I}_{\mathrm{img}}$ and text $\mathbf{T}_{\mathrm{txt}}$ are then encoded separately to obtain visual features $\bigl[\mathbf{f}_{\mathrm{cls}}, \mathbf{F}_v\bigr]$, instruction embeddings $\mathbf{f}_{\mathrm{ins}}$, and textual features $\bigl[\mathbf{t}_{\mathrm{cls}}, \mathbf{F}_t\bigr]$. Next, we compute the belief matrix $\mathbf{M}^{\mathrm{bel}}_{j}$ between visual features and instruction embeddings, and calculate its rank $R_{j}$ in batch training. Each element $\mathbf{F}_{:,l}$ of $\bigl[\mathbf{f}_{\mathrm{cls}}, \mathbf{F}_v\bigr] \in \mathbb{R}^{d \times (m+1)}$ is then refined via rank-weighted aggregation to obtain $\mathbf{F}_v^{(\text{soft})}$. Finally, Spatial-PAE and Temporal-PAE are applied to capture long-range dependencies, yielding the final visual embedding $\mathbf{v}_{\mathrm{emb}}$ and text embedding $\mathbf{t}_{\mathrm{emb}}$, respectively. This design provides an unbiased vision representation and strengthens textual encoding through cyclic activation.  

For open-domain retrieval, we propose a two-stage prior representation learning strategy. In the first stage, the model is pre-trained on the large-scale RS5M dataset~\cite{10679571} with contrastive loss to learn generalizable vision–language representations. In the second stage, fine-tuning is performed on RSITR datasets, following a procedure similar to the closed-domain case but with two key distinctions. (i) The visual backbone is updated based on base ViT and ResNet models, while (ii) the text encoder is replaced with a language transformer trained on a large-scale remote sensing corpus, eliminating the need for an additional Spatial-PAE. Moreover, PriorCLIP further strengthens the ViT architecture by incorporating convolutional patching, a soft-belief refinement strategy, and a transformer encoder to enhance interactions between visual and instructional features. 
  
In summary, PriorCLIP introduces (1) a soft-belief strategy to mitigate semantic noise, (2) Spatial-PAE and Temporal-PAE for robust closed-domain retrieval, and (3) a two-stage prior learning scheme with architecture enhancements for open-domain retrieval, collectively leading to improved adaptability and retrieval performance across both settings.

\begin{table*}[t]
\renewcommand\arraystretch{1.15}
\caption{Comparison of closed-domain image-text retrieval results on RSICD and RSITMD datasets. 
Best results are \textbf{bold}, second best are \underline{underlined}.}
\label{tab:table1}
\centering
\resizebox{\linewidth}{!}{
\begin{tabular}{lcc|ccc|ccc}
\toprule
\multirow{3}{*}{\textbf{Method}} & \multirow{3}{*}{\textbf{Params}} & 
\multirow{3}{*}{\begin{tabular}[c]{@{}c@{}}\textbf{Backbone} \\ vision / text\end{tabular}} &
\multicolumn{3}{c|}{\textbf{RSICD Dataset}} & 
\multicolumn{3}{c}{\textbf{RSITMD Dataset}} \\
\cmidrule(lr){4-6} \cmidrule(lr){7-9}
& & & 
\begin{tabular}[c]{@{}c@{}}Image$\to$Text \\ R@1 / R@5 / R@10\end{tabular} & 
\begin{tabular}[c]{@{}c@{}}Text$\to$Image \\ R@1 / R@5 / R@10\end{tabular} & 
mR &
\begin{tabular}[c]{@{}c@{}}Image$\to$Text \\ R@1 / R@5 / R@10\end{tabular} & 
\begin{tabular}[c]{@{}c@{}}Text$\to$Image \\ R@1 / R@5 / R@10\end{tabular} & 
mR \\ 
\midrule
$VSE_0$ \cite{faghri2017vse++} (TL)   & 29M  & ResNet-50 / GRU & 4.56 / 16.73 / 22.94 & 4.37 / 15.37 / 25.35 & 14.89 & 9.07 / 21.61 / 31.78 & 7.73 / 27.80 / 41.00 & 23.17 \\
SCAN i2t \cite{lee2018stacked}       & 60M  & RoI Trans / GRU & 4.82 / 13.66 / 21.99 & 3.93 / 15.20 / 25.53 & 14.19 & 8.92 / 22.12 / 33.78 & 7.43 / 25.71 / 39.03 & 22.83 \\
SCAN t2i \cite{lee2018stacked}       & 60M  & RoI Trans / GRU & 4.79 / 16.19 / 24.86 & 3.82 / 15.70 / 28.28 & 15.61 & 7.01 / 20.58 / 30.90 & 7.06 / 26.49 / 42.21 & 22.37 \\ 
CAMP \cite{yuan2021lightweight}     & 63M  & RoI Trans / GRU & 4.64 / 14.61 / 24.09 & 4.25 / 15.82 / 27.82 & 15.20 & 8.11 / 23.67 / 34.07 & 6.24 / 26.37 / 42.37 & 23.47 \\
CAMERA \cite{qu2020context}         & 64M  & RoI Trans / BERT & 4.57 / 13.08 / 21.77 & 4.00 / 15.93 / 26.97 & 14.39 & 8.33 / 21.83 / 33.11 & 7.52 / 26.19 / 40.72 & 22.95 \\ 
\midrule
LW-MCR \cite{yuan2021lightweight}   & -    & CNNs / CNNs      & 3.29 / 12.52 / 19.93 & 4.66 / 17.51 / 30.02 & 14.66 & 10.18 / 28.98 / 39.82 & 7.79 / 30.18 / 49.78 & 27.79 \\
AMFMN \cite{yuan2022exploring}      & 36M  & ResNet-18 / GRU  & 5.21 / 14.72 / 21.57 & 4.08 / 17.00 / 30.60 & 15.53 & 10.63 / 24.78 / 41.81 & 11.51 / 34.69 / 54.87 & 29.72 \\
GaLR \cite{yuan2022remote}          & 46M  & ResNet-18+ppyolo / GRU & 6.59 / 19.85 / 31.04 & 4.69 / 19.48 / 32.13 & 18.96 & 14.82 / 31.64 / 42.48 & 11.15 / 36.68 / 51.68 & 31.41 \\
KCR \cite{mi2022knowledge}          & -    & ResNet-101 / BERT & 5.95 / 18.59 / 29.58 & 5.40 / 22.44 / 37.36 & 19.89 & - & - & - \\
SWAN \cite{pan2023reducing}         & 40M  & ResNet-50 / GRU   & 7.41 / 20.13 / 30.86 & 5.56 / 22.26 / 37.41 & 20.61 & 13.35 / 32.15 / 46.90 & 11.24 / 40.40 / 60.60 & 34.11 \\ 
HVSA \cite{zhang2023hypersphere}    & -    & ResNet-50 / GRU   & 7.47 / 20.62 / 32.11 & 5.51 / 21.13 / 34.13 & 20.61 & 13.20 / 32.08 / 45.58 & 11.43 / 39.20 / 57.45 & 33.16 \\ 
DOVE \cite{ma2023direction}         & 38M  & ResNet-50+RoI Trans / GRU & 8.66 / 22.35 / 34.95 & 6.04 / 23.95 / 40.35 & 22.72 & 16.81 / 36.80 / 49.93 & 12.20 / \textbf{49.93} / \textbf{66.50} & 37.73 \\ 
\midrule
$VSE_1$ \cite{faghri2017vse++} (CL)  & 131M & ViT-B / BERT & 9.06 / 22.78 / 32.75 & 5.32 / 19.47 / 33.71 & 20.52 & 12.83 / 31.19 / 46.24 & 9.60 / 36.59 / 54.42 & 31.81 \\
$VSE_2$ \cite{faghri2017vse++} (TL)  & 137M & Swin-T / BERT & 7.96 / 21.68 / 34.31 & 6.44 / 22.09 / 37.49 & 21.66 & 17.70 / 36.28 / 48.67 & \textbf{13.58} / 41.24 / 59.29 & 36.13 \\
$VSE_2$ \cite{faghri2017vse++} (CL)  & 137M & Swin-T / BERT & \underline{10.52} / 25.07 / 36.78 & 6.11 / 23.64 / 38.28 & 23.40 & 16.15 / 40.04 / \underline{53.10} & 10.75 / 38.98 / 60.18 & 36.53 \\
$VSE_2$ \cite{faghri2017vse++} $+$ $TEL$ (CL) & 161M & Swin-T / BERT & 10.43 / 25.89 / 37.15 & 6.02 / 23.42 / 38.30 & 23.53 & 17.48 / 36.95 / 50.44 & 11.81 / 41.99 / 61.77 & 36.74 \\ 
\midrule
\rowcolor{blue!10}\textbf{PriorCLIP-CD (hard)} & 161M & Swin-T+ResNet-50 / BERT & 9.88 / \textbf{27.26} / \textbf{39.16} & \underline{6.97} / \underline{24.56} / \underline{38.92} & \underline{24.46} & \underline{18.14} / \underline{41.15} / 52.88 & 12.17 / 41.68 / \underline{63.41} & \underline{38.24} \\ 
\rowcolor{blue!10}\textbf{PriorCLIP-CD (soft)} & 161M & Swin-T+ResNet-50 / BERT & \textbf{10.89} / \underline{26.17} / \underline{37.79} & \textbf{7.17} / \textbf{25.07} / \textbf{41.06} & \textbf{24.69} & \textbf{18.36} / \textbf{42.04} / \textbf{55.53} & \underline{13.36} / \underline{44.47} / 61.73 & \textbf{39.25} \\ 
\bottomrule
\end{tabular}}
\end{table*}

\subsection{Loss Function}\label{Se3.5}
\subsubsection{\textbf{Contrastive Loss}}
{In RSITR, triplet loss \cite{karpathy2015deep} is commonly used for model optimization. However, contrastive loss provides better optimization performance, as shown in the comparison between $VSE_2$ (TL) and $VSE_2$ (CL) in Tab. \ref{tab:table1}. Therefore, we adopt contrastive loss \cite{hadsell2006dimensionality} to optimize our model, as it requires only pairwise similarity calculations without the need for additional negative samples.}

Given the $\mathbf{v}_{emb}$ and $\mathbf{t}_{e m b}$ embedded by the PriorCLIP from batch image-text pairs, we calculate the cosine similarity between them as $\bm{s}_{i,j}=\mathbf{v}_{i}^{\mathrm{T}}\mathbf{t}_{j}$. Then we can get the contrastive loss as
\begin{align}
\mathcal{L}_c = -\frac{1}{N} \sum_{i=1}^{N} \left[ \log \frac{\exp\left(\bm{s}_{i,i} / \tau \right)}{\sum_{j=1}^{N} \exp \left( \bm{s}_{i,j} / \tau \right)} \right. \notag \\
+ \left. \log \frac{\exp\left(\bm{s}_{i,i} / \tau \right)}{\sum_{j=1}^{N} \exp \left( \bm{s}_{j,i} / \tau \right)} \right],
\end{align}
where $\tau$ is the temperature parameter.

\subsubsection{\textbf{Affiliation Loss}}
Most existing retrieval tasks rely primarily on pair-wise objectives such as contrastive loss~\cite{radford2021learning,zeng2021multi} or triplet loss~\cite{karpathy2015deep}. While effective, these objectives treat samples independently and overlook the underlying distributional structure of the data (see Fig.~\ref{fig:fig5}(a)). This often results in semantic confusion zones, where samples from different categories are not well separated.  

To address this limitation, we propose an Affiliation Loss, which explicitly exploits the distributional characteristics of remote sensing data by minimizing the distance between each modality and the cluster center of its affiliated category~\cite{pan2023reducing}, as illustrated in Fig.~\ref{fig:fig5}(b).  

Affiliation Loss is a cluster-based objective~\cite{li2021shapenet} that integrates contrastive learning with scene category supervision to enforce tighter intra-class compactness and clearer inter-class separability. As shown in Alg.~\ref{alg:batch_cluster_loss}, samples in a mini-batch are first grouped into $C$ categories based on scene-level annotations\footnote{For RSICD and RSITMD, category information can be obtained directly from the image file names.}. Each image and its paired text are assigned to the same category, yielding positive pairs $\left\{(\mathbf{v}_i, \mathbf{t}_i^*) \mid i=1,2,\dots, N\right\}$, where $\mathbf{t}_i^*$ denotes the cluster center of the $i$-th text. Symmetrically, we define vision-centered clusters for text samples.  

\begin{table*}[t]
\renewcommand\arraystretch{1.15}
\caption{Comparison of open-domain image-text retrieval results on RSICD and RSITMD datasets. 
$\dagger$ denotes zero-shot CLIP and $\ddagger$ denotes fine-tuned CLIP. 
Best results are \textbf{bold}, second best are \underline{underlined}.}
\label{tab:table9}
\centering
\resizebox{\linewidth}{!}{
\begin{tabular}{lcc|cc|cc|c}
\toprule
\multicolumn{8}{c}{\textbf{RSICD Dataset}} \\
\midrule
\textbf{Method} & \textbf{Params} & \begin{tabular}[c]{@{}c@{}}Backbone \\ vision / text\end{tabular} & 
\textbf{Pre-train} & \textbf{Fine-tune} & 
\begin{tabular}[c]{@{}c@{}}Image$\to$Text \\ R@1 / R@5 / R@10\end{tabular} & 
\begin{tabular}[c]{@{}c@{}}Text$\to$Image \\ R@1 / R@5 / R@10\end{tabular} & 
\textbf{mR} \\
\midrule
CLIP $\dagger$ \cite{radford2021learning} & 151M & ViT-B / Transformer & WIT (CLIP) & - & 4.76 / 12.81 / 19.12 & 4.70 / 15.43 / 25.01 & 13.64 \\
CLIP-FT $\ddagger$ \cite{radford2021learning} & 151M & ViT-B / Transformer & WIT (CLIP) & RSICD & 15.55 / 30.56 / 41.99 & 14.68 / 33.63 / 42.71 & 29.85 \\
PE-RSITR $\ddagger$ \cite{yuan2023parameter} & 151M & ViT-B / Transformer & WIT (CLIP) & RSICD & 14.13 / 31.51 / 44.78 & 11.63 / 33.92 / 50.73 & 31.12 \\
RemoteCLIP $\dagger$ \cite{remoteclip} & 151M & ViT-B / Transformer & RET-3+DET-10+SEG-4 & - & 17.02 / 37.97 / 51.51 & 13.71 / 37.11 / 54.25 & 35.26 \\
RemoteCLIP $\dagger$ \cite{remoteclip} & 428M & ViT-L / Transformer & RET-3+DET-10+SEG-4 & - & 18.39 / 37.42 / 51.05 & 14.73 / 39.93 / 56.58 & 36.35 \\
SkyCLIP $\dagger$ \cite{Wang_Prabha_Huang_Wu_Rajagopal_2024} & 151M & ViT-B / Transformer & Skyscript & - & 6.59 / 16.10 / 26.53 & 7.14 / 22.34 / 34.29 & 18.83 \\
SkyCLIP $\dagger$ \cite{Wang_Prabha_Huang_Wu_Rajagopal_2024} & 428M & ViT-L / Transformer & Skyscript & - & 7.04 / 17.75 / 27.08 & 6.40 / 21.21 / 34.00 & 18.91 \\
GeoRSCLIP $\dagger$ \cite{10679571} & 151M & ViT-B / Transformer & RS5M & - & 11.53 / 28.55 / 39.16 & 9.52 / 27.37 / 40.99 & 26.18 \\
GeoRSCLIP-FT $\ddagger$ \cite{10679571} & 151M & ViT-B / Transformer & RS5M & RSICD & 22.14 / 40.53 / 51.78 & 15.26 / 40.46 / \underline{57.79} & 38.00 \\
GeoRSCLIP-FT $\ddagger$ \cite{10679571} & 151M & ViT-B / Transformer & RS5M & RET-2 & 21.13 / 41.72 / \textbf{55.63} & 15.59 / 41.19 / \textbf{57.99} & 38.87 \\
\midrule
\rowcolor{blue!10}\textbf{PriorCLIP-OD $\ddagger$} & 189M & ResNet-50+ViT-B / Transformer & RS5M & RSICD & \underline{26.62} / 40.71 / 50.32 & \textbf{21.35} / \underline{43.59} / 54.27 & 39.48 \\
\rowcolor{blue!10}\textbf{PriorCLIP-OD $\ddagger$} & 189M & ResNet-50+ViT-B / Transformer & RS5M & RET-2 & 25.62 / \underline{43.18} / 53.06 & 19.91 / 42.74 / 55.15 & \underline{39.95} \\
\rowcolor{blue!10}\textbf{PriorCLIP-OD $\ddagger$} & 189M & ResNet-50+ViT-B / Transformer & RS5M & RET-3 & \textbf{27.63} / \textbf{45.38} / \underline{55.26} & \underline{21.10} / \textbf{44.87} / 56.12 & \textbf{41.73} \\
\midrule
\multicolumn{8}{c}{\textbf{RSITMD Dataset}} \\
\midrule
CLIP $\dagger$ \cite{radford2021learning} & 151M & ViT-B / Transformer & WIT (CLIP) & - & 5.31 / 17.26 / 25.00 & 5.84 / 22.74 / 34.82 & 18.50 \\
CLIP-FT $\ddagger$ \cite{radford2021learning} & 151M & ViT-B / Transformer & WIT (CLIP) & RSITMD & 25.00 / 48.23 / 62.17 & 23.72 / 49.38 / 63.63 & 45.35 \\
PE-RSITR $\ddagger$ \cite{yuan2023parameter} & 151M & ViT-B / Transformer & WIT (CLIP) & RSITMD & 23.67 / 44.07 / 60.36 & 20.10 / 50.63 / 67.97 & 44.47 \\
RemoteCLIP $\dagger$ \cite{remoteclip} & 151M & ViT-B / Transformer & RET-3+DET-10+SEG-4 & - & 27.88 / 50.66 / 65.71 & 22.17 / 56.46 / 73.41 & 49.38 \\
RemoteCLIP $\dagger$ \cite{remoteclip} & 428M & ViT-L / Transformer & RET-3+DET-10+SEG-4 & - & 28.76 / 52.43 / 63.94 & 23.76 / 59.51 / 74.73 & 50.52 \\
SkyCLIP $\dagger$ \cite{Wang_Prabha_Huang_Wu_Rajagopal_2024} & 151M & ViT-B / Transformer & Skyscript & - & 10.18 / 25.44 / 35.62 & 10.88 / 33.27 / 49.82 & 27.54 \\
SkyCLIP $\dagger$ \cite{Wang_Prabha_Huang_Wu_Rajagopal_2024} & 428M & ViT-L / Transformer & Skyscript & - & 12.61 / 28.76 / 38.05 & 10.62 / 34.73 / 51.42 & 29.37 \\
GeoRSCLIP $\dagger$ \cite{10679571} & 151M & ViT-B / Transformer & RS5M & - & 19.03 / 34.51 / 46.46 & 14.16 / 42.39 / 57.52 & 35.68 \\
GeoRSCLIP-FT $\ddagger$ \cite{10679571} & 151M & ViT-B / Transformer & RS5M & RSITMD & 30.09 / 51.55 / 63.27 & 23.54 / \underline{57.52} / \underline{74.60} & 50.10 \\
GeoRSCLIP-FT $\ddagger$ \cite{10679571} & 151M & ViT-B / Transformer & RS5M & RET-2 & 32.30 / 53.32 / 67.92 & 25.04 / \textbf{57.88} / 74.38 & 51.81 \\
\midrule
\rowcolor{blue!10}\textbf{PriorCLIP-OD $\ddagger$} & 189M & ResNet-50+ViT-B / Transformer & RS5M & RSITMD & 30.97 / 58.19 / 71.02 & 25.62 / 55.27 / \textbf{75.00} & 52.68 \\
\rowcolor{blue!10}\textbf{PriorCLIP-OD $\ddagger$} & 189M & ResNet-50+ViT-B / Transformer & RS5M & RET-2 & \underline{44.25} / \textbf{65.71} / \textbf{75.22} & \underline{28.45} / 54.65 / 68.05 & \underline{56.05} \\
\rowcolor{blue!10}\textbf{PriorCLIP-OD $\ddagger$} & 189M & ResNet-50+ViT-B / Transformer & RS5M & RET-3 & \textbf{45.58} / \underline{65.49} / \underline{75.00} & \textbf{30.13} / 55.44 / 68.54 & \textbf{56.70} \\
\bottomrule
\end{tabular}}
\end{table*}

The vision-to-text and text-to-vision cosine similarities are computed as $\bm{s}_{i,j}^v=\mathbf{v}_{i}^{\mathrm{T}}\mathbf{t}_{j}^*$ and $\bm{s}_{j,i}^t=\mathbf{t}_{j}^{\mathrm{T}}\mathbf{v}_{i}^*$, respectively. The cluster-based symmetric contrastive loss is then formulated as:  
\begin{align}
\mathcal{L}_a = -\frac{1}{N} \sum_{i=1}^{N} \Bigg[ 
\log \frac{\exp \left(\bm{s}_{i,i}^v / \tau\right)}{\sum_{j=1}^{N} \exp \left(\bm{s}_{i,j}^v / \tau\right)} \notag \\
+ \log \frac{\exp \left(\bm{s}_{i,i}^t / \tau\right)}{\sum_{j=1}^{N} \exp \left(\bm{s}_{j,i}^t / \tau\right)} 
\Bigg],
\end{align}
where $\tau$ is a temperature scaling factor. The final optimization objective combines the conventional contrastive loss $\mathcal{L}_c$ with our proposed affiliation loss:  
\begin{equation}
\mathcal{L}_{total} = \mathcal{L}_{c} + \bm{\lambda}_{cs}\mathcal{L}_{a},
\end{equation}
where $\bm{\lambda}_{cs}$ is a balancing coefficient controlling the influence of category clustering. Affiliation Loss goes beyond pair-wise alignment by leveraging class-level structure, effectively reducing semantic confusion zones and yielding more discriminative representations for remote sensing retrieval.

\section{Experiments}
\subsection{Experimental Settings}

\subsubsection{\textbf{RSICD Dataset}}  
The RSICD dataset comprises 10,921 remote sensing images with a resolution of $224 \times 224$, each paired with five corresponding textual captions. Following the protocol in~\cite{yuan2022exploring,yuan2022remote}, we split the dataset into 7,862 training images, 1,966 validation images, and 1,093 test images.  

\subsubsection{\textbf{RSITMD Dataset}}  
The RSITMD dataset contains 4,743 images with a resolution of $256 \times 256$, each annotated with five captions that are more fine-grained than those in RSICD. Following~\cite{yuan2022remote}, we adopt the official split of 3,435 training images, 856 validation images, and 452 test images.  

\subsubsection{\textbf{RS5M Dataset}}  
The RS5M dataset~\cite{10679571} comprises approximately 5 million remote sensing image-text pairs, collected by filtering publicly available resources and automatically captioning label-only datasets using pre-trained image captioning models. In this work, RS5M serves as the large-scale pre-training corpus for PriorCLIP.  

\subsubsection{\textbf{Evaluation Metrics}}  
In line with prior RSITR studies~\cite{yuan2021lightweight,yuan2022exploring}, we adopt R@K ($K=1, 5, 10$) and mean recall (mR) as evaluation metrics. R@K measures the percentage of correctly retrieved matches within the top-$K$ results, while mR computes the average of all R@K values, offering a comprehensive assessment of retrieval performance.

\begin{table}[t]
\renewcommand\arraystretch{1.15}
\caption{Ablation on the RSITMD test set. $\dagger$ denotes a modified S-PAE. 
Best results are \textbf{bold}, second best are \underline{underlined}.}
\label{tab:table2}
\centering
\resizebox{\linewidth}{!}{
\begin{tabular}{lcc|c}
\toprule
\textbf{Method} &
\begin{tabular}[c]{@{}c@{}}Image$\to$Text \\ R@1 / R@5 / R@10\end{tabular} &
\begin{tabular}[c]{@{}c@{}}Text$\to$Image \\ R@1 / R@5 / R@10\end{tabular} &
\textbf{mR} \\
\midrule
\multicolumn{4}{c}{\textbf{PriorCLIP for Closed-Domain RSITR}} \\
\midrule
\textit{baseline}                   & 16.15 / 40.04 / 53.10 & 10.75 / 38.98 / 60.18 & 36.53 \\
\textit{+S-PAE}                     & 16.59 / 38.50 / 54.65 & 12.88 / 40.04 / 58.19 & 36.81 \\
\textit{+T-PAE}                     & \textbf{20.13} / 41.15 / 53.10 & \textbf{14.60} / 38.89 / 53.81 & 36.95 \\
\textit{+$\mathcal{L}_{a}$}         & \underline{19.25} / 36.95 / 52.88 & 10.88 / 39.51 / 59.87 & 36.56 \\
\textit{+S-PAE+T-PAE}               & 17.04 / 39.16 / \underline{54.42} & \underline{13.50} / 40.04 / 59.34 & 37.25 \\
\textit{+S-PAE+$\mathcal{L}_{a}$}   & 16.81 / \underline{40.04} / \textbf{55.31} & 11.59 / 41.50 / \underline{61.02} & \underline{37.71} \\
\textit{+T-PAE+$\mathcal{L}_{a}$}   & 17.04 / 38.72 / 53.76 & 11.81 / \underline{41.64} / 60.58 & 37.26 \\
\rowcolor{blue!10}\textit{+S-PAE+T-PAE+$\mathcal{L}_{a}$} 
                                     & 18.14 / \textbf{41.15} / 52.88 & 12.17 / \textbf{41.68} / \textbf{63.41} & \textbf{38.24} \\
\midrule
\multicolumn{4}{c}{\textbf{PriorCLIP for Open-Domain RSITR}} \\
\midrule
\textit{CLIP-baseline}              & 25.00 / 48.23 / 62.17 & 23.72 / 49.38 / 63.63 & 45.35 \\
\textit{+S-PAE $\dagger$}           & 24.56 / \underline{53.32} / \underline{66.37} & 21.02 / 51.86 / \underline{71.02} & 48.02 \\
\textit{+$\mathcal{L}_{a}$}         & \underline{27.65} / 52.21 / 63.94 & \underline{25.49} / \underline{53.10} / 66.59 & \underline{48.16} \\
\rowcolor{blue!10}\textit{+S-PAE $\dagger$ + $\mathcal{L}_{a}$} 
                                     & \textbf{30.97} / \textbf{58.19} / \textbf{71.02} & \textbf{25.62} / \textbf{55.27} / \textbf{75.00} & \textbf{52.68} \\
\bottomrule
\end{tabular}}
\end{table}

\subsection{Implementation Details}
For closed-domain retrieval, we adopt the Swin Transformer (Swin-T, tiny version)~\cite{krizhevsky2017imagenet} as the vision encoder, pre-trained on ImageNet, and BERT (base version) as the text encoder, utilizing the official pre-trained parameters. Both vision and text features are projected from 768 to 512 dimensions through a linear layer. Self-attention and cross-attention modules are configured with eight heads and a dropout rate of 0.2, while the number of stacked units is set to 2 for Spatial-PAE and 3 for Temporal-PAE. For instruction encoding, we employ ResNet-50, pre-trained on the AID dataset~\cite{xia2017aid}, using the second-to-last layer (dimension 1024) as the output, which is then linearly mapped to 512 dimensions. The contrastive loss and affiliation loss are optimized with a temperature coefficient of 0.07 and a center factor of 1, respectively.  

For open-domain retrieval, we employ ResNet-50 + ViT-B as the visual backbone and a Transformer as the textual backbone, aligning with the vanilla CLIP approach. A modified ResNet-50 is pre-trained on AID~\cite{xia2017aid}, and parameters before the attention pooling layer are frozen during CLIP tuning. Training is conducted for 20 epochs in pre-training (with a batch size of 256) and 10 epochs in fine-tuning (with a batch size of 512), each comprising 1,000,000 training steps per epoch. The source code and pre-trained model checkpoints are publicly available \footnote{\url{https://github.com/jaychempan/PriorCLIP}.}.

\subsection{Performance Comparisons}
\subsubsection{\textbf{State-of-the-art Methods}}  
We compare our method with both closed-domain and open-domain retrieval approaches.  

\textbf{Closed-Domain RSITR.}  
These baselines are grouped as follows:  
\begin{itemize}
  \item \textbf{Generic image-text retrieval models:} $VSE_0$~\cite{faghri2017vse++}, SCAN~\cite{lee2018stacked}, CAMP~\cite{yuan2021lightweight}, and CAMERA~\cite{qu2020context}.  
  \item \textbf{Remote sensing-specific models:} LW-MCR~\cite{yuan2021lightweight}, AMFMN~\cite{yuan2022exploring}, GaLR~\cite{yuan2022remote}, KCR~\cite{mi2022knowledge}, SWAN~\cite{pan2023reducing}, HVSA~\cite{zhang2023hypersphere}, and DOVE~\cite{ma2023direction}.  
  \item \textbf{Transformer-based variants:} $VSE_1$, $VSE_2$~\cite{faghri2017vse++}, and $VSE_2$ $+$ $TEL$.  
\end{itemize}

\begin{table}[t]
\renewcommand\arraystretch{1.15}
\caption{Effect of filter size in Spatial-PAE on the RSITMD test set. 
Best results are \textbf{bold}, second best are \underline{underlined}.}
\label{tab:table3}
\centering
\resizebox{\linewidth}{!}{
\begin{tabular}{lcc|c}
\toprule
\textbf{Filter Size ($\bm{r}_k$)} &
\begin{tabular}[c]{@{}c@{}}Image$\to$Text \\ R@1 / R@5 / R@10\end{tabular} &
\begin{tabular}[c]{@{}c@{}}Text$\to$Image \\ R@1 / R@5 / R@10\end{tabular} &
\textbf{mR} \\
\midrule
$10$ (hard) & 17.26 / 40.71 / \underline{54.65} & 13.01 / 41.06 / 61.19 & 37.98 \\
$20$ (hard) & 17.48 / 39.60 / 52.21 & \textbf{13.50} / 42.48 / \underline{62.74} & 38.00 \\
$30$ (hard) & \textbf{19.03} / 37.83 / 49.78 & 12.83 / 41.90 / 62.08 & 37.24 \\
$40$ (hard) & 18.14 / \underline{41.15} / 52.88 & 12.17 / 41.68 / \textbf{63.41} & \underline{38.24} \\
$50$ (hard) & 18.36 / 37.61 / 50.88 & 13.10 / \underline{42.88} / 62.04 & 37.48 \\
\midrule
\rowcolor{blue!10}$50$ (soft) & \underline{18.36} / \textbf{42.04} / \textbf{55.53} & \underline{13.36} / \textbf{44.47} / 61.73 & \textbf{39.25} \\
\bottomrule
\end{tabular}}
\end{table}

For SCAN, CAMP, and CAMERA, we follow prior work and adopt RoI Trans~\cite{ding2019learning} with ResNet-50 pre-trained on DOTA~\cite{xia2018dota} to detect salient objects. For remote sensing-specific models, we report the best results from the original papers, as they are difficult to reproduce on RSICD and RSITMD. To ensure fairness, we further extend VSE baselines with Transformer backbones, where $VSE_i$ ($i=0,1,2$) denotes different encoder configurations and TEL indicates an additional Transformer encoder layer. Loss functions are annotated as TL (triplet loss) or CL (contrastive loss).  

Finally, we include two closed-domain variants of our method: ``PriorCLIP-CD (hard)'' and ``PriorCLIP-CD (soft)'', which apply hard- and soft-belief strategies, respectively.  

\textbf{Open-Domain RSITR.}  
These approaches train a visual-semantic embedding model with a dual-tower structure on large-scale image-text datasets using natural language supervision. Representative models include:  

\begin{itemize}
  \item \textbf{CLIP}~\cite{radford2021learning}, trained on 400M web image-text pairs (denoted as WIT (CLIP));  
  \item \textbf{RemoteCLIP}~\cite{remoteclip}, pre-trained on a mixture of remote sensing datasets, including RET-3 (three retrieval datasets: RSICD, RSITMD, UCM-Captions~\cite{10.1145/1869790.1869829}), DET-10 (ten detection datasets), and SEG-4 (four segmentation datasets);  
  \item \textbf{GeoRSCLIP}~\cite{10679571}, trained on RS5M, a large-scale remote sensing image-text corpus;  
  \item \textbf{SkyCLIP}~\cite{Wang_Prabha_Huang_Wu_Rajagopal_2024}, trained on the SkyScript dataset.  
\end{itemize}

RET-3 is constructed by combining three retrieval datasets and further de-duplicated following  RemoteCLIP~\cite{remoteclip} to prevent data leakage. In addition, we construct RET-2 (RSICD + RSITMD), which is also de-duplicated, as an alternative fine-tuning set for GeoRSCLIP.  

For our method in open-domain retrieval, we pre-train PriorCLIP-OD on RS5M~\cite{10679571}, which, to the best of our knowledge, is the largest publicly available remote sensing image-text dataset. For fairness, the results of CLIP and SkyCLIP are obtained using officially released weights, whereas the results of RemoteCLIP and GeoRSCLIP are reported from their original papers.

\subsubsection{\textbf{Quantitative Comparison of Closed-Domain Methods}}  
Closed-domain experiments are conducted under the setting where training and testing are performed on the same dataset, as summarized in Tab.~\ref{tab:table1}.

\begin{figure*}[t]
  \centering
  \includegraphics[width=0.85\linewidth]{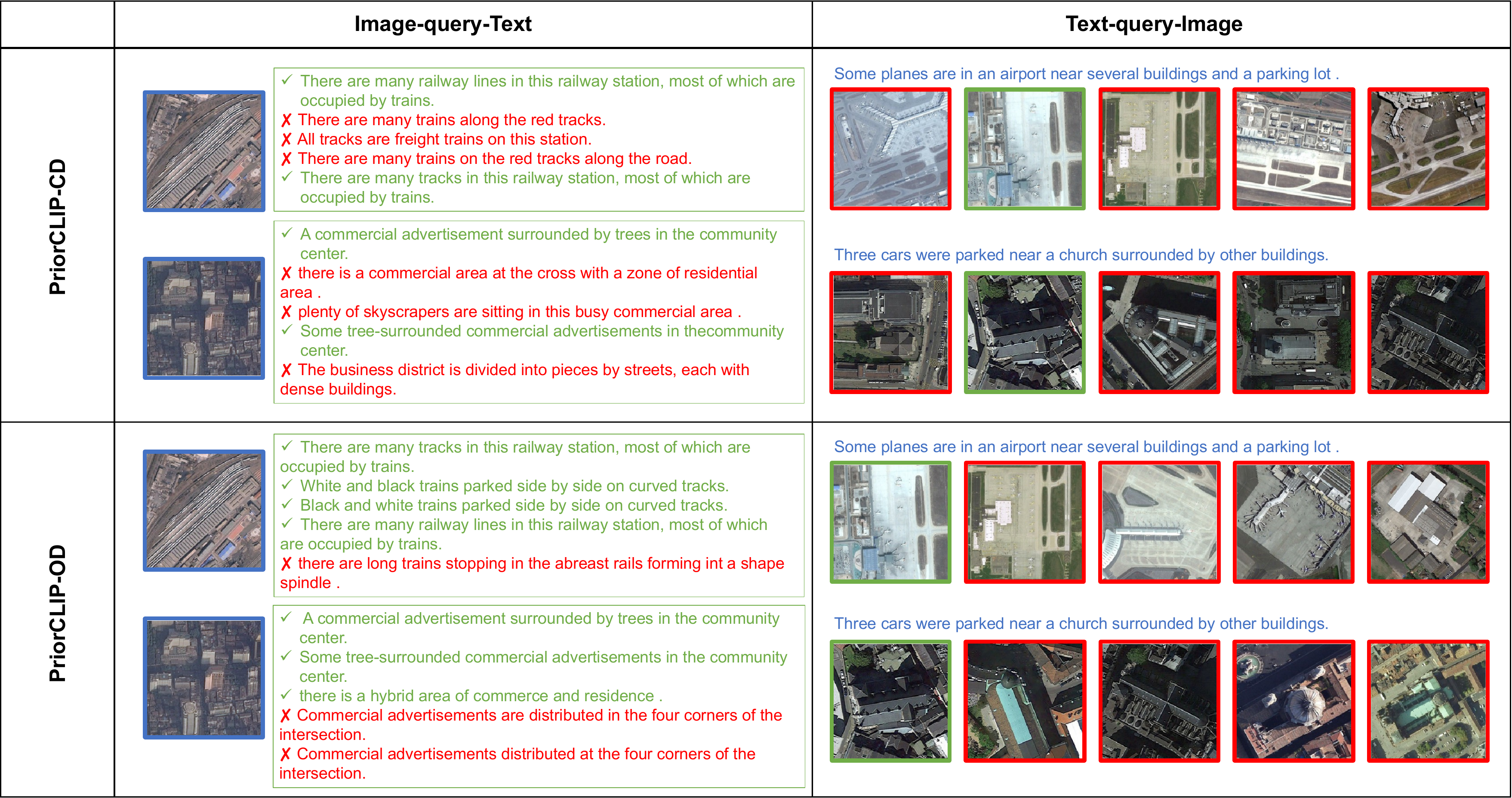}
  \caption{The visualization of the top 5 retrieval results includes Image-query-Text (left) and Text-query-Image (right). Blue text and boxes represent queries, green text and boxes represent matching retrieval results, and red text and boxes indicate mismatched retrieval results.}
  \label{fig:fig6}
\end{figure*}

\textbf{Results on RSICD.}  
As presented in Tab.~\ref{tab:table1}, remote sensing-specific approaches consistently outperform generic image-text retrieval baselines. Within Transformer-based encoders, the Swin Transformer~\cite{liu2021swin} provides more robust visual features than the ViT~\cite{dosovitskiy2020image}. A comparison between $VSE_2$ trained with triplet loss and contrastive loss further indicates that contrastive loss is more effective. Retrieval performance also varies significantly depending on the backbone configuration: models with Swin-T combined with BERT perform best, followed by ViT with BERT, while ResNet-50 combined with GRU yields the weakest results. This demonstrates the advantage of Transformer-based architectures over CNN-based ones for RSITR tasks.  

The addition of a Transformer encoder layer enables $VSE_2$ $+$ $TEL$ to achieve an mR of 23.53, although the improvement over $VSE_2$ remains limited. In contrast, our proposed PriorCLIP substantially enhances retrieval performance, achieving an mR of 24.69, which represents a 4.9\% relative improvement over the strongest baseline.  

\textbf{Results on RSITMD.}  
Compared with RSICD, the RSITMD dataset contains more fine-grained textual descriptions, resulting in overall higher retrieval performance. As shown in Tab.~\ref{tab:table1}, traditional image-text retrieval methods remain unsatisfactory across both datasets. Among remote sensing-specific models, SWAN, which combines CNN with GRU, achieves an mR of 34.11. Transformer-based methods provide further improvements, with $VSE_2$ $+$ $TEL$ reaching an mR of 36.74.  

Our PriorCLIP delivers the best performance, attaining an mR of 39.24, which corresponds to a 4.0\% relative improvement over the strongest baseline. Moreover, PriorCLIP achieves R@1 scores of 18.36\% for image-to-text (I2T) and 13.36\% for text-to-image (T2I).  

\subsubsection{\textbf{Quantitative Comparison of Open-Domain Methods}}  
Open-domain experiments involve fine-tuning on small-scale RSITR datasets after pre-training on additional large-scale corpora, as summarized in Tab.~\ref{tab:table9}.

\textbf{Results on RSICD.}  
The zero-shot performance of CLIP varies considerably depending on the pre-training corpus. For example, RemoteCLIP achieves an mR of 36.35 on RSICD. Results from RemoteCLIP and SkyCLIP further indicate that scaling the backbone from ViT-B to ViT-L yields only marginal gains. Fine-tuning the WIT-trained CLIP model improves mR from 13.64 to 29.85 on RSICD, and further fine-tuning on RSITR datasets yields additional improvements. GeoRSCLIP, fine-tuned on RET-2, achieves an mR of 38.87.  

Our PriorCLIP, fine-tuned on RSICD, RET-2, and RET-3, achieves additional improvements of 1.5\%, 2.7\%, and 7.3\%, respectively, over the GeoRSCLIP baseline. Notably, PriorCLIP also delivers the strongest R@1 gains, with 27.63\% for I2T and 21.35\% for T2I. These results demonstrate the effectiveness of our two-stage CLIP-based framework in enhancing both image and text retrieval.  

\begin{figure*}[th]
  \centering
  \includegraphics[width=0.85\linewidth]{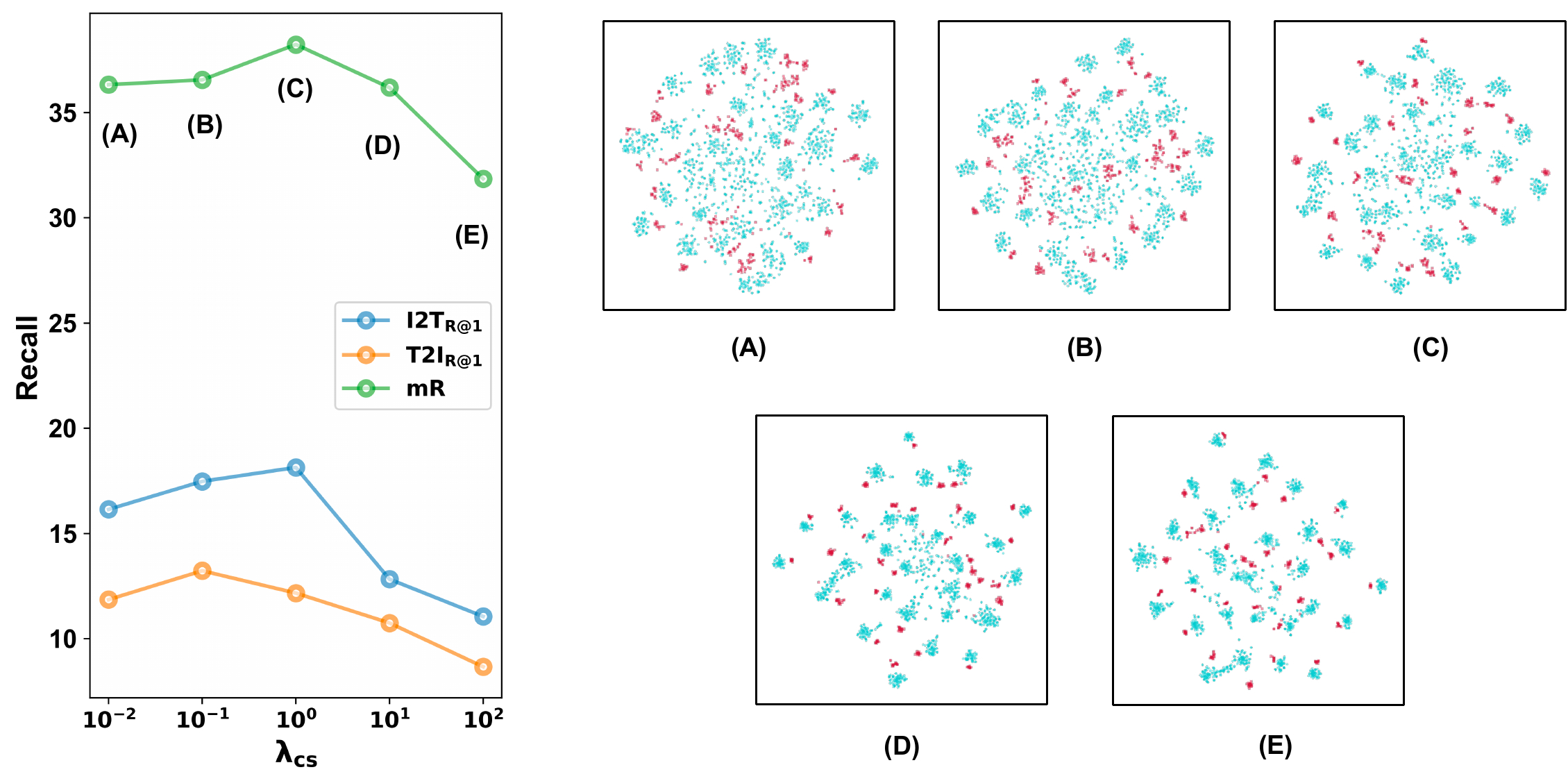}
  \caption{The effect of the center scale on retrieval results (left) and RSITMD test set visualized with t-SNE \cite{van2008visualizing} in the embedding space (right), the red squares represent vision embeddings, and the blue triangles represent text embeddings.}
  \label{fig:fig7}
\end{figure*}

\textbf{Results on RSITMD.}  
On RSITMD, RemoteCLIP achieves an mR of 50.52 in the zero-shot setting. Fine-tuning WIT-based CLIP substantially improves performance, raising mR from 18.50 to 45.35. Compared with RemoteCLIP (ViT-L) pre-trained on RET-3, PriorCLIP achieves a 9.6\% higher mR despite using fewer parameters. GeoRSCLIP fine-tuned on RET-2 further increases mR performance to 51.81.  

Our PriorCLIP surpasses this benchmark, yielding additional mR gains of 1.7\%, 8.2\%, and 9.4\% when fine-tuned on RSITMD, RET-2, and RET-3, respectively. In terms of R@1, PriorCLIP achieves 45.58\% for I2T and 30.13\% for T2I. These results highlight the benefits of our vision-instruction fine-tuning strategy in strengthening visual representations and improving open-domain retrieval performance.  

\subsubsection{\textbf{Visualization of Closed-Domain and Open-Domain Methods}}  
Fig.~\ref{fig:fig6} is qualitative examples of the top-5 retrieval results for both \texttt{Image-query-Text} (left) and \texttt{Text-query-Image} (right). In \texttt{Image-query-Text}, the blue box denotes the query image, green text indicates correctly matched captions, and red text corresponds to mismatched results. In \texttt{Text-query-Image}, the blue text denotes the query sentence, while green and red boxes represent correct and incorrect image matches, respectively.

For \texttt{Image-query-Text}, most of the top-5 retrieved captions correctly correspond to the query image, despite each image being associated with only five ground-truth captions. For \texttt{Text-query-Image}, the correct image is typically ranked first or second among the retrieved results, even though each caption is paired with only a single ground-truth image.  

Overall, PriorCLIP provides more semantically coherent retrievals and assigns higher ranks to correct matches compared with competing methods. In \texttt{Text-query-Image}, although some retrieved images are visually similar but semantically less precise, the ranking gap between correct and incorrect results is larger under PriorCLIP, suggesting that our model effectively suppresses misleading candidates. Nonetheless, certain failure cases remain, where semantically related but incorrect results are retrieved, highlighting opportunities for further refinement.

\subsection{Ablation Studies}  
We conduct twelve ablation experiments (eight for PriorCLIP-CD and four for PriorCLIP-OD) to evaluate the contribution of three key components: Spatial-PAE, Temporal-PAE, and the proposed affiliation loss $\mathcal{L}_{a}$, as summarized in Tab.~\ref{tab:table2}. The \textit{baseline} setting removes all three modules, while \textit{CLIP-baseline} denotes PriorCLIP without Spatial-PAE and affiliation loss, keeping all other settings unchanged.  

\subsubsection{\textbf{Effects of Vision Instruction Representation}}  
The addition of Spatial-PAE consistently improves retrieval performance. Compared with the \textit{baseline}, \textit{+S-PAE} yields a modest increase in mR, accompanied by a clear gain in T2I R@1. When combined with Temporal-PAE, Spatial-PAE further raises mR (36.95 $\rightarrow$ 37.25), with notable improvements in I2T and T2I R@10. Incorporating affiliation loss amplifies these benefits: \textit{+S-PAE+T-PAE+$\mathcal{L}_{a}$} outperforms \textit{+T-PAE+$\mathcal{L}_{a}$} by 2.6\% in mR, with I2T R@1 increasing from 17.04 to 18.14 and T2I R@1 from 11.81 to 12.17.  

For PriorCLIP-OD, Spatial-PAE improves mR from 45.35 to 48.02 relative to \textit{CLIP-baseline}, and from 48.16 to 52.68 when combined with $\mathcal{L}_{a}$. These results confirm that Spatial-PAE significantly strengthens vision representation and enhances T2I retrieval.  

\subsubsection{\textbf{Effects of Language Cycle Attention}}  
Temporal-PAE provides substantial improvements in language representation. Relative to the \textit{baseline}, \textit{+T-PAE} increases I2T R@1 and T2I R@1 by 24.6\% and 35.8\%, respectively. Adding Temporal-PAE to Spatial-PAE yields an additional 1.2\% improvement in mR, while incorporating it into \textit{+S-PAE+$\mathcal{L}_{a}$} results in a further 1.4\% mR gain (I2T R@1: 16.81 $\rightarrow$ 18.14; T2I R@1: 11.59 $\rightarrow$ 12.17).  

Consistent improvements across multiple configurations demonstrate that Temporal-PAE complements Spatial-PAE and strengthens retrieval in both I2T and T2I tasks.  

\subsubsection{\textbf{Effects of Affiliation Loss}}  
The affiliation loss $\mathcal{L}_{a}$ also contributes significantly. Compared with the \textit{baseline}, \textit{+$\mathcal{L}_{a}$} increases I2T R@1 from 16.15 to 19.25. When applied together with Spatial-PAE and Temporal-PAE, $\mathcal{L}_{a}$ further raises mR by 2.7\% (I2T R@1: 17.04 $\rightarrow$ 18.14).  

In the open-domain setting, $\mathcal{L}_{a}$ improves mR from 45.35 to 48.16 relative to \textit{CLIP-baseline}, and from 48.02 to 52.68 when combined with Spatial-PAE. These results indicate that affiliation loss not only enhances vision and language representations but also enables Spatial-PAE and Temporal-PAE to capture long-range dependencies better.

\subsection{Parameter Evaluation}  

\subsubsection{\textbf{Filter Size in PriorCLIP}}  
In PriorCLIP with the hard-belief strategy, the visual features extracted by the Transformer often contain semantic noise. In contrast, only a subset of these features is highly representative of the underlying visual semantics. To mitigate the effect of noise, we regulate the number of key features by adjusting the filter size parameter. The Vision Transformer outputs a maximum of fifty local features, from which a subset is selected.  

Tab.~\ref{tab:table3} reports the bidirectional retrieval performance on the RSITMD dataset under different filter sizes. When the filter size is set to ten, image-to-text retrieval achieves the highest R@10 of 54.65. At a filter size of twenty, mean Recall reaches 38.00, while text-to-image R@1 attains its maximum of 13.50. A filter size of thirty again benefits I2T R10 but yields relatively lower mR. The best balance is observed when the filter size is forty, achieving the highest mR of 38.24, suggesting that a moderate number of features captures the most representative visual cues. When the filter size is expanded to fifty, mR declines to 37.48, indicating that including all features reintroduces semantic noise that suppresses the most informative ones.  

In contrast, PriorCLIP with the soft-belief strategy avoids the need to fix the number of features. It dynamically assigns weights, reaching an mR of 39.25, which outperforms all fixed filter sizes. These findings highlight that semantic noise in remote sensing images tends to deactivate key features, and performance can be enhanced by reducing reliance on redundant or noisy visual features.  

\subsubsection{\textbf{Center Scale of Affiliation Loss}}  
The affiliation loss is designed to reduce semantic confusion in the shared embedding space by encouraging intra-class compactness. The strength of this constraint is governed by the center scale parameter, denoted as $\lambda_{cs}$. The left panel of Fig.~\ref{fig:fig7} shows retrieval performance on RSITMD with varying center scales, while the right panel illustrates the corresponding embedding distributions.  

As the center scale increases, the inter-class separation becomes more pronounced, and semantic confusion zones gradually diminish, as illustrated by regions A to E in Fig.~\ref{fig:fig7}. However, retrieval accuracy first improves and then declines. The optimal performance is achieved when $\lambda_{cs}=10^0$, which balances intra-class compactness with inter-class separation. Extreme constraints may lead to over-clustering, thereby limiting generalization and reducing retrieval effectiveness.

\subsection{Instruction Strategy Analysis}  
To investigate the role of instruction encoders, we evaluate PriorCLIP with different encoder architectures under the hard-belief strategy. Each encoder is pre-trained on classification tasks using the AID~\cite{xia2017aid}, RESISC45~\cite{7891544}, or ImageNet~\cite{deng2009imagenet} datasets. The candidate encoders are divided into two categories: (i) CNN-based architectures, including VGG-16~\cite{simonyan2014very}, VGG-19~\cite{simonyan2014very}, ResNet-50~\cite{he2016deep}, and ResNet-101~\cite{he2016deep}; and (ii) Transformer-based architectures, ViT~\cite{dosovitskiy2020image} and Swin-T~\cite{liu2021swin}.  

\begin{figure}[t]
  \centering
  \includegraphics[width=\linewidth]{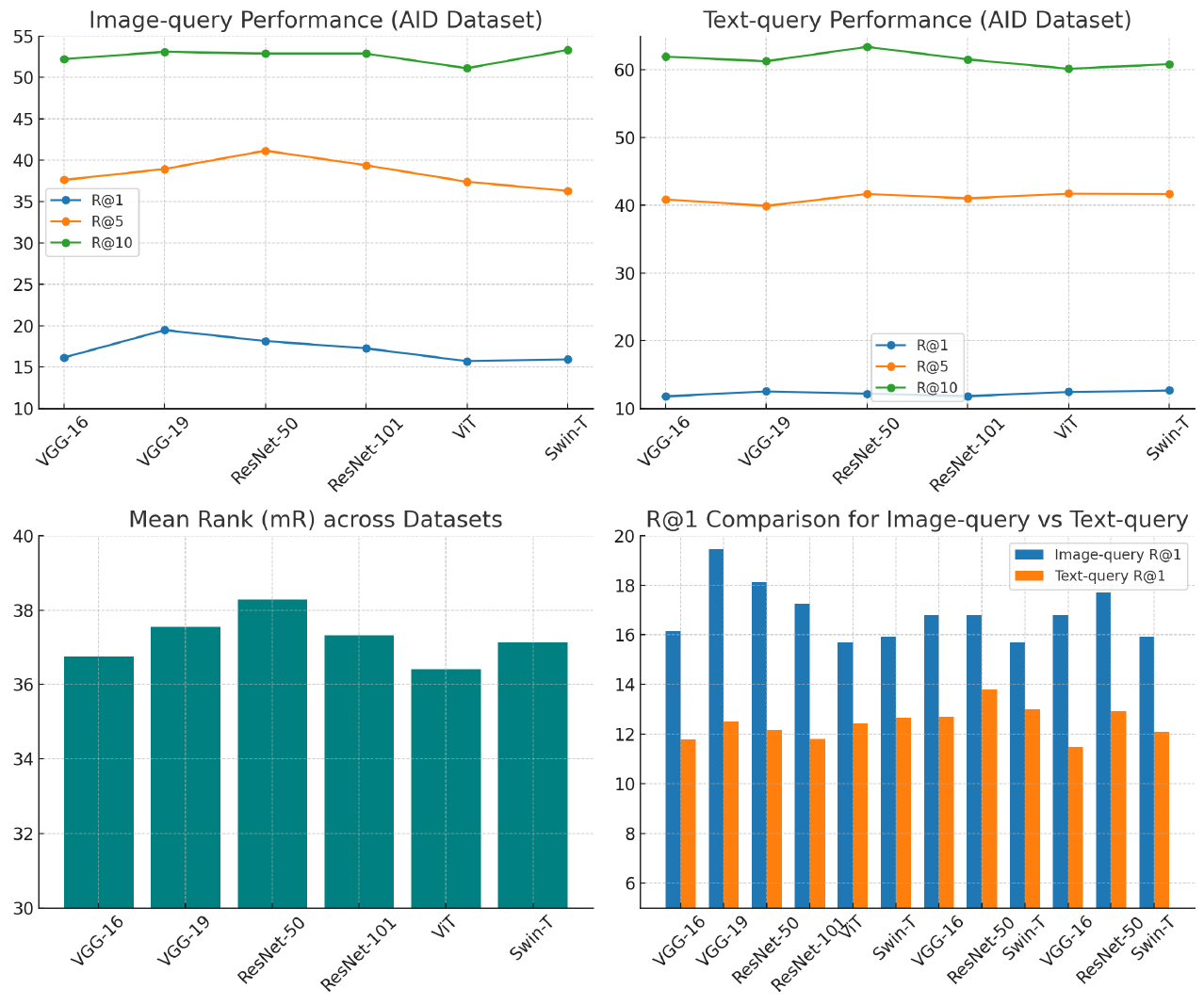}
  \caption{Results of experiments of different instruction strategies on the RSITMD test set.}
  \label{fig:figx}
\end{figure}

Experimental results on the RSITMD dataset are presented in Fig.~\ref{fig:figx}. Among all configurations, ResNet-50 consistently achieves the highest performance, with mean Recall (mR) values of 38.24 and 38.28 when pre-trained on AID and RESISC45, respectively. In contrast, Transformer-based encoders such as ViT and Swin-T provide no significant improvements when pre-trained on AID, suggesting that Transformer-based priors are less effective for this task. A comparison between VGG-16 and ResNet-50 trained on ImageNet further indicates that pre-training on remote sensing scene classification datasets provides stronger and more relevant prior knowledge than pre-training on generic natural images.  

Overall, these findings demonstrate that instruction strategies derived from remote sensing scene recognition deliver more effective priors for guiding visual representations. Leveraging such domain-specific knowledge enables PriorCLIP to produce more reliable embeddings and, consequently, improved retrieval performance.  

\section{CONCLUSION}
In this work, we proposed the visual prior-guided vision-language model, a novel paradigm that leverages prior knowledge from the remote sensing domain to enable adaptive learning of vision and text representations. Within this model, we developed PriorCLIP for remote sensing image-text retrieval. Specifically, Spatial-PAE and Temporal-PAE were designed to model long-range dependencies and enhance key feature representations. For vision representation, prior knowledge from remote sensing scene recognition was utilized to construct unbiased features and suppress semantic noise. For text representation, Temporal-PAE recurrently activated the current time step using the previous step, thereby strengthening semantic encoding.
Furthermore, we introduced a vision-instruction fine-tuning strategy to mitigate semantic noise in remote sensing vision--language representations, enabling more robust open-domain retrieval. In addition, a cluster-wise Attribution Loss was proposed to enforce inter-class constraints and reduce semantic confusion in the shared subspace. Extensive experiments on RSICD and RSITMD verified the superiority and effectiveness of PriorCLIP, consistently achieving state-of-the-art performance in both closed-domain and open-domain scenarios.

\bibliographystyle{IEEEtran}

\bibliography{references}
\end{document}